%% file: main.tex
\documentclass[11pt,twoside]{article}
\usepackage{fullpage}

\usepackage{epsf}
\usepackage{fancyhdr}
\usepackage{graphics}
\usepackage{graphicx}
\usepackage{psfrag}
\usepackage{microtype}
\usepackage{algorithmic}
\usepackage{color,xcolor}
\usepackage{caption}
\usepackage{subcaption}
\usepackage[linesnumbered,ruled]{algorithm2e}
\DontPrintSemicolon
\usepackage{color}

\usepackage{amsthm}
\usepackage{amsfonts}
\usepackage{amsmath}
\usepackage{amssymb,bbm}

\usepackage{mathrsfs}

\usepackage{url}
\usepackage[colorlinks,linkcolor=magenta,citecolor=blue, pagebackref=true,backref=true]{hyperref}
\renewcommand*{\backrefalt}[4]{%
    \ifcase #1 \footnotesize{(Not cited.)}%
    \or        \footnotesize{(Cited on page~#2.)}%
    \else      \footnotesize{(Cited on pages~#2.)}%
    \fi}

\usepackage{nicefrac}
\usepackage{comment}
\usepackage{soul}
\usepackage{chngpage}

 \usepackage{tabularx}%
\usepackage{enumitem}
\usepackage{booktabs}
\usepackage{caption}

\usepackage{bm,bbm}
\usepackage{mathtools}
\usepackage{cleveref}

\usepackage{eucal}

\usepackage{scalerel}

\input{final_macros}

\newcommand{\strongconvex}{\mu}
\newcommand{\smooth}{L}

\newtheorem{assumption}{Assumption}
\newcommand{\simiid}{\stackrel{\mathrm{i.i.d.}}{\sim}}
\newcommand{\numobs}{\ensuremath{n}}

\newcommand{\usedim}{\ensuremath{d}}

\newenvironment{carlist}
 {\begin{list}{$\bullet$}
 {\setlength{\topsep}{0in} \setlength{\partopsep}{0in}
  \setlength{\parsep}{0in} \setlength{\itemsep}{\parskip}
  \setlength{\leftmargin}{0.07in} \setlength{\rightmargin}{0.08in}
  \setlength{\listparindent}{0in} \setlength{\labelwidth}{0.08in}
  \setlength{\labelsep}{0.1in} \setlength{\itemindent}{0in}}}
 {\end{list}}

\newcommand{\filtration}{\mathcal{F}}

\newcommand{\fakerefassumelip}[1]{\hyperref[assume:smooth-high-order]{{\color{magenta} {\upshape\textbf (}{\upshape{\textbf{Lip}}#1}{\upshape\textbf )}}} }

\DeclareFontFamily{U}{mathx}{}
\DeclareFontShape{U}{mathx}{m}{n}{<-> mathx10}{}
\DeclareSymbolFont{mathx}{U}{mathx}{m}{n}

\DeclareMathAccent{\widecheck}{0}{mathx}{"71}

\usepackage{mathbbol}

\long\def\comment#1{}

\usepackage[framemethod=TikZ]{mdframed}

\newenvironment{narrowpara}
  {\par\addvspace{\smallskipamount}\narrower\noindent\ignorespaces}
  {\par\addvspace{\smallskipamount}}

\usepackage{wrapfig}
\newcommand{\coordinate}{e}

\newcommand{\sigstarj}[1]{\sigma_{*, #1}}
\newcommand{\sigstar}{\sigma_*}

\newcommand{\correctionMap}{\mathcal{H}}

\newcommand{\DSet}{\mathcal{D}}

\newcommand{\sglip}{\ell}

\newcommand{\smallscale}{\omega}

\newcommand{\channel}{\mathcal{C}}
\newcommand{\quantizer}{\mathcal{Q}}

\newcommand{\qtzalg}{\quantizer_{\mathrm{D}}}
\newcommand{\qtzadc}{\quantizer_{\mathrm{C}}}

\newcommand{\signalup}{g}
\newcommand{\signaldown}{h}

\begin{document}

\begin{center}
{\bf{\LARGE{Federated learning over physical channels: adaptive algorithms with near-optimal guarantees}}}

\vspace*{.2in}
{\large{
 \begin{tabular}{cc}
  Rui Zhang$^{ \dagger, \star}$ 
 \end{tabular}
 \begin{tabular}{cc}
  Wenlong Mou$^{ \diamond, \star}$ 
 \end{tabular}

}

\vspace*{.2in}

 \begin{tabular}{c}
 Department of Electrical Engineering, SUNY University at Buffalo$^{\dagger}$
 \end{tabular}

 \begin{tabular}{c}
 Department of Statistical Sciences, University of Toronto$^{\diamond}$
 \end{tabular}

}

\begin{abstract}
  In federated learning, communication cost can be significantly reduced by transmitting the information over the air through physical channels. In this paper, we propose a new class of adaptive federated stochastic gradient descent (SGD) algorithms that can be implemented over physical channels, taking into account both channel noise and hardware constraints. We establish theoretical guarantees for the proposed algorithms, demonstrating convergence rates that are adaptive to the stochastic gradient noise level. We also demonstrate the practical effectiveness of our algorithms through simulation studies with deep learning models.
\let\thefootnote\relax\footnote{$^\star$RZ and WM contributed equally to this work.}
\end{abstract}
\end{center}

\section{Introduction}
In modern machine learning applications, large datasets are often distributed across multiple heterogeneous worker machines. To jointly solve the optimization problem, the distributed machines need to transmit the information through communication channels. The bottleneck of distributed and federated learning is often the communication cost~\cite{li2014communication,li2020federated}. The development of efficient federated learning algorithms with low communication cost has been a central topic in the machine learning community for the past decade (see~\cite{zhang2013communication,boyd2011distributed,chang2020distributed,mcmahan2017communication} and references therein).

Most federated learning literature focuses on network-layer abstraction of communication channels, which allows error-free transmissions of real-valued data~\cite{arjevani2015communication}. Nevertheless, such a transmission scheme can be costly in practice, requiring error correction codes and high-precision floating-point numbers. On the other hand, first-order stochastic optimization algorithms are known to be resilient to random noises in gradient oracles. In particular, if we want to minimize an objective function $F$, given a stochastic gradient oracle $\widehat{g}(\theta)$ satisfying the conditions
\begin{align}
  \Exs \big[ \widehat{g} (\theta) \mid \theta \big] = \nabla F (\theta), \quad \mbox{and} \quad \Exs \big[ \vecnorm{\widehat{g} (\theta)}{2}^2 \mid \theta \big] < + \infty,\label{eq:oracle-condition-intro}
\end{align}
various stochastic first-order methods are guaranteed to converge efficiently. This observation motivates study of federated learning over physical communication channels~\cite{amiri2020machine,yang2020federated}. For example, when transmitting stochastic gradients directly through analogue channels with Additive Gaussian White Noise (AWGN), the stochastic gradient oracle $\widehat{g}(\theta)$ satisfies the condition~\eqref{eq:oracle-condition-intro}, thereby achieving the same convergence guarantees with significantly reduced communication costs.~\cite{amiri2020federated,amiri2020federatedq,amiri2021convergence,wei2022federated,xiao2024over}.

Despite the inspiring progress, existing federated learning algorithms over physical channels suffer from both practical limitations and theoretical gaps. From communication channel perspective, existing works overlook hardware constraints in practical systems: due to the non-linear conversion mapping between analogue and digital signals, \Cref{eq:oracle-condition-intro} is not satisfied in general, and the biases in stochastic oracle may deteriorate the convergence of stochastic optimization algorithms. From the optimization perspective, existing algorithms either require fully coded communication in one of the uplink and the downlink~\cite{amiri2020machine,amiri2021convergence}, or require the noise level to decay sufficiently fast~\cite{wei2022federated,upadhyay2023noisy} -- under these transmission schemes, the reduction in communication costs are limited. Finally, the noisy communication channel may amplify the variance of stochastic gradient oracles, leading to sub-optimal performance compared to coded channels. These limitations motivate the key question:
\begin{quote}
  Can we significantly reduce the communication costs of federated learning using practical physical channels, while retaining desirable performance guarantees?
\end{quote}
We answer this question in the affirmative by introducing a new class of channels signal processing techniques and adaptive stochastic gradient descent (SGD) algorithms for federated learning over physical channels. Our contribution are threefold:
\begin{itemize}
  \item To tackle the biases induced by analogue-to-digital conversion (ADC), we introduce a stochastic post-coding procedure that corrects the biases in the quantized signals. The post-coding procedure is adaptive to the noise level in the communication channel, and can be implemented with low computational overhead.
  \item Under a worker-server architecture, we propose a new class of adaptive federated SGD algorithms that transmit the stochastic gradient information primarily through physical channels, and use the coded channel to synchronize parameters only once in a while. The communication cost is significantly lower than existing algorithms.
  \item We establish theoretical guarantees for the proposed algorithms, demonstrating near-optimal convergence rates. In particular, we show that the error bounds are adaptive to the stochastic gradient noise level, achieving statistical risks comparable to those of coded channels.
\end{itemize}
The rest of the paper is organized as follows. We first introduce notations and discuss related work. The problem setup is presented in \Cref{sec:problem-setup}. We present the main algorithm in \Cref{sec:main-algorithm}, and the theoretical guarantees in \Cref{sec:theory}. The results are validated through simulation studies in \Cref{sec:simulation}, and through theoretical proofs in \Cref{sec:proofs}. We conclude the paper with a discussion about future work in \Cref{sec:discussion}.

\paragraph{Notations:} We use $[m]$ to denotes the set $\{1,2,\ldots,m\}$. For $j \in [\usedim]$, we use $\coordinate_j \in \real^\usedim$ to denote an indicator vector with $1$ in the $j$-th coordinate and $0$ elsewhere. We use $f \circ g$ to denote the composition of functions $f$ and $g$, i.e., $(f \circ g)(x) = f(g(x))$. For $p \in [1, + \infty)$, we denote the vector $\ell^p$ norm by $\vecnorm{x}{p} \mydefn \big( \sum_{j=1}^{\usedim} |x_j|^p \big)^{1/p}$, and $\vecnorm{x}{\infty} \mydefn \max_{j \in [\usedim]} |x_j|$. Given a pair of vector norms $\vecnorm{\cdot}{X}$ and $\vecnorm{\cdot}{Y}$, we use $\matsnorm{A}{X \to Y} \mydefn \sup_{\vecnorm{x}{X} = 1} \vecnorm{Ax}{Y}$ to denote the induced matrix norm. We use $\inprod{x}{y}$ to denote the standard inner product in $\real^\usedim$. For the iterative algorithms we study we use $\filtration_k$ to denote the $\sigma$-field generated by the first $k$ iterations.

\subsection{Additional related works}
\label{subsec:related-works}
Recently, federated learning over noisy channels has gained significant research attention. In addition to aforementioned works, several works have explored the interplay between communication and learning in this context. Extending the basic AWGN channel model, several practical communication scenarios are considered, including fading channels~\cite{shah2022robust,amiri2020federated}, multi-path effect~\cite{tegin2021blind}, heavy-tailed interference~\cite{li2025robust}, and quantization~\cite{amiri2020federated,tegin2021blind}. Note that the stochastic noise becomes biased after applying the non-linear quantization mapping, creating obstacles for convergence of gradient-based algorithms. To our knowledge, our work is the first to construct an exact unbiased gradient oracle under this setting.

The performance of federated learning relies on estimation error for the gradient information on the receiver side. Advanced signal processing techniques have been employed to reduce the estimation error.
\cite{tegin2021blind} studies OFDM channels and shows that the noise can be mitigated by increasing number of receiver antennae. \cite{zhang2022coded} combines channel coding techniques with the noisy transmission to reduce the error. \cite{ang2020robust} used regularization to improve the robustness of federated learning under noisy channels. Additionally, a recent line of research~\cite{guo2022joint,yang2022over,yao2024wireless} studied resource allocation strategies for wireless federated learning, focusing on optimizing the trade-off between communication efficiency and learning performance. 

\Cref{sec:main-algorithm} will present three main techniques we introduce to ensure near-optimal performance. Let us discuss connection of these techniques with the existing literature.
\begin{carlist}
  \item The post-coding procedure is related to dithering~\cite{wannamaker2002theory}, which injects random perturbations before quantization to reduce error; similar ideas appear in distributed computation~\cite{aysal2008distributed} and federated learning~\cite{hasirciouglu2024communication}.
  \item The scale-adaptive transformation, which separates scale and normalized value, is akin to techniques for efficient low-precision training~\cite{koster2017flexpoint,gupta2015deep}.
  \item Periodic synchronization of global model parameters at the server is a classical strategy in distributed optimization~\cite{stich2018local,karimireddy2020scaffold}, here used to reduce coded communication rounds. Combining this with further communication reduction is an interesting future direction.
\end{carlist}

\section{Problem setup}\label{sec:problem-setup}
We consider a federated optimization problem with $m$ workers machines. Each worker machine $j \in \{1,2, \cdots, m\}$ is associated with a probability distribution $\Prob_j$ and a dataset $\DSet_j = \big( X_i^{(j)} \big)_{i = 1}^{\numobs}$, such that $X_i^{(j)} \simiid \Prob_j$ for each $j$. Our goal is to jointly solve the following optimization problem
\begin{align}
  \min_{\theta \in \real^\usedim} F (\theta) \mydefn \Exs_\Prob \big[ f (\theta; X) \big] \quad \mbox{where } \Prob = \frac{1}{m} \sum_{j = 1}^m \Prob_j.\label{eq:opt-objective}
\end{align}
A central server machine is used to aggregate information from different worker machines. In particular, a communication link exists between each worker machine $j \in [m]$ and the server machine. The algorithm can choose to transmit information through either coded or physical channels. In the following subsection, we will discuss these two types of channels in detail.

\subsection{Models for physical constraints}\label{subsec:physical-constraints}
In this section, we summarize the physical models for communication channels and hardware devices considered in this paper.

\subsubsection{Coded vs. physical transmissions channels}
In standard coded communication systems, the gradients and model parameters are transmitted as floating numbers through a coded channel. To transmit a real number with a floating number precision of $2^{-b}$, we need $b$ bits to encode the information. The information is further modulated as PAM signal. Given a PAM of order $2^\ell$, and an error correction code with overhead rate $\alpha$, the average number of symbols needed to transmit a real number is $\frac{b}{\ell} (1 + \alpha)$. For example, with a 32-bit floating precision number, PAM-4 modulation, and $20\%$ overhead, we need $9.6$ symbols on average to transmit a real number.

Over-the-air communication, on the other hand, transmits the real numbers as analogue signals directly. Due to the random noise in the communication channels and the hardware constraints, information cannot be transmitted exactly in such channels. Throughout this paper, we consider a simple channel with Additive Gaussian White Noise (AWGN). Given an input sequence $(X_1, X_2, \cdots, X_t)$, the channel $\channel$ outputs
\begin{align}
  Y_i = X_i + \varepsilon_i, \quad \mbox{for $\varepsilon_i \simiid \mathcal{N} (0, \sigma_c^2)$}.\label{eq:awgn-channel}
\end{align}
In addition to the AWGN model, we also consider the impact of hardware constraints on the communication process.

\subsubsection{Conversion between analogue and digital signals}
The information is stored as floating point numbers in the memory of digital devices. To transmit the information through physical channels, the floating point numbers need to be converted to analogue signals through a digital-to-analogue converter (DAC). On the other hand, the received analogue signals are converted back to digital signals through an analogue-to-digital converter (ADC).

Concretely, let the quantization levels be $z_1< z_2< \cdots< z_q$ with $|z_i - z_{i - 1}| = \Delta$ for each $i$, the DAC hardware takes digital representation for one of the levels $z_i$ as input, and the output is the corresponding analogue signal. The ADC hardware takes the received analogue signal $x$ as input and outputs its nearest quantized level, i.e., the ADC component implements a deterministic mapping defined by
\begin{align*}
  \qtzadc (x) = \arg\min_{z_i} |x - z_i|.
\end{align*}
Note that the numerical precision of floating point numbers is usually much higher than that of the quantized levels, i.e., we have $|z_i - z_{i + 1}| \gg 2^{-b}$. This means that the quantization process can introduce significant errors, especially when the input signal $x$ is not well-aligned with the quantization levels. In order to mitigate the quantization error, we use a randomized algorithmic quantizer $\qtzalg$ that maps floating-point data to the quantization levels, before passing through the DAC.
\begin{align}
   \qtzalg (x) = \begin{cases}
    z_1 & \mbox{if } x < z_1,\\
    z_q & \mbox{if } x \geq z_q,\\
    z_{i + \iota} & \mbox{if } x \in [z_i, z_{i + 1}), \mbox{ where }\iota \sim \mathrm{Ber} \Big( \frac{x - z_i}{z_{i + 1} - z_i} \Big).
   \end{cases}\label{eq:random-qtz}
\end{align}
The randomized mapping makes the algorithmic quantizer $\qtzalg$ unbiased, i.e., for any $x \in [-1, 1]$, we have $\Exs [ \qtzalg (x)] = x$. This randomized quantization scheme has been employed in federated learning literature~\cite{amiri2020federatedq,youn2023randomized}.

The output of the algorithmic quantizer is then transmitted through the DAC, the channel, and the ADC, sequentially. The overall mapping from the real-valued data to the received digital signals is given by $\qtzadc \circ \channel \circ \qtzalg$.

\section{Adaptive SGD algorithms for physical channels}\label{sec:main-algorithm}
Let us now present the main algorithm and theoretical guarantees. The communication between workers and the centralized server follows the following protocol: assuming that a two-way link has been established between the server and each worker, they can exchange information through two types of channels:
\begin{itemize}
  \item A physical channel, where the signals pass through the DAC unit, the channel, and the ADC unit,sequentially. The channel takes a quantized scalar value $x \in \{ z_1, z_2, \cdots, z_q\}$ as input, and outputs $\qtzadc \circ \channel (x)$ at the receiver side.
  \item A coded channel, which takes bit sequences as input, and uses error correction codes to guarantee error-free transmissions.
\end{itemize}
Although the additive Gaussian noises in physical channels are unbiased, when combined with the nonlinear quantization process, they can lead to biased estimates, which jeopardize the convergence of stochastic optimization algorithms. To address this issue, we introduce a stochastic post-coding procedure to correct the bias, as described in the following section.

\subsection{Stochastic post-coding}\label{subsec:post-coding}
For $i \in [q]$, the composition mapping $\qtzadc \circ \channel$ is generally biased, i.e., $\Exs [ \qtzadc \circ \channel (z_i)] \neq z_i$. The goal of the stochastic post-coding procedure is to construct a stochastic mapping $\correctionMap$, such that
\begin{align}
  \Exs \big[ \correctionMap \circ \qtzadc \circ \channel (z_i) \big] = z_i, \quad \mbox{for each } i \in \{2,3,\cdots, q - 1\}.\label{eq:post-coding-unbiasedness}
\end{align}
Note that we only guarantee the unbiasedness of the mapping for the quantization levels in the interior of the quantization grid. Since the output space is constrained in $\{z_1, z_2, \cdots, z_q\}$, the mapping $\correctionMap \circ \qtzadc \circ \channel$ cannot guarantee unbiasedness for the boundary points $z_1$ and $z_q$. However, as long as $q \geq 4$, we can still carry information using only the interior points. Let us now describe the post-coding mapping $\correctionMap$.

To set up the problem, we use $P$ to indicate the transition probabilities of the mapping $\qtzadc \circ \channel$, i.e., for $i, j \in [q]$, we define
\begin{align*}
  P_{i,j} = \Prob \big( \qtzadc \circ \channel (z_i) = z_j \big) = \begin{cases}
    \Phi \big( \frac{z_j + \Delta / 2 - z_i}{\sigma_c} \big) - \Phi \big( \frac{z_j - \Delta / 2 - z_i}{\sigma_c} \big), & j \in \{2,3,\cdots, q - 1\},\\
    \Phi \big( \frac{z_1 + \Delta / 2 - z_i}{\sigma_c} \big), & j = 1,\\
    1 - \Phi \big( \frac{z_q - \Delta / 2 - z_i}{\sigma_c} \big), & j = q,
  \end{cases}
\end{align*}
where $\Phi$ is the CDF of the standard normal distribution.
\begin{subequations}\label{eq:lp-for-postcode}
We use the transition matrix $H$ to represent the mapping $\correctionMap$, i.e., $H_{i,j} = \Prob \big( \correctionMap (z_i) = z_j \big)$. We solve the following linear program to find the matrix $H$:
\begin{align}
 &\min_{H \in \real^{q \times q}, v \in \real} ~ v \label{eq:lp-for-postcode-objective}\\
 \mbox{s.t.}\quad & H_{i,j} \geq 0, \quad \forall i,j \in [q];\qquad \sum_{j = 1}^q H_{i,j} = 1, \quad \forall i \in [q]; \label{eq:lp-for-postcode-constraint-1}\\
 & \coordinate_j^\top P H z = z_j \quad \forall j \in \{2,3,\cdots, q - 1\}, \label{eq:lp-for-postcode-constraint-2} \\
 & \sum_{i = 1}^q (PH)_{j,i} \cdot (z_i - z_j)^2 \leq v \quad \forall j \in \{2,3,\cdots, q - 1\}. \label{eq:lp-for-postcode-constraint-3}
\end{align}
\end{subequations}
The constraint~\eqref{eq:lp-for-postcode-constraint-1} ensures that the mapping $\correctionMap$ is a valid probability transition matrix, and the constraint \eqref{eq:lp-for-postcode-constraint-2} guarantees the unbiasedness property~\eqref{eq:post-coding-unbiasedness}. Under the constraint~\eqref{eq:lp-for-postcode-constraint-3}, the objective function~\eqref{eq:lp-for-postcode-objective} minimizes the worst-case variance of the mapping $\correctionMap \circ \qtzadc \circ \channel$. The following fact is clear from our construction.
\begin{proposition}\label{prop:basic-post-coding-property}
  If the linear program~\eqref{eq:lp-for-postcode} is feasible, with an optimal solution $(H^*, v^*)$. Then the mapping $\correctionMap$ satisfies the unbiasedness property~\eqref{eq:post-coding-unbiasedness}, and
  \begin{align*}
    \var \big( \correctionMap \circ \qtzadc \circ \channel (z_i) \big) \leq v^*, \quad \forall i \in \{2,3,\cdots, q - 1\}.
  \end{align*}
\end{proposition}
The construction of the mapping $\correctionMap$ relies on the feasibility of the linear program~\eqref{eq:lp-for-postcode}. While feasibility is not guaranteed in general, the following lemma shows that the linear program is feasible when the SNR is sufficiently large.

\begin{lemma}\label{lemma:feasibility-post-coding}
  For any $\sigma_c \leq \Delta / 2$, the linear program~\eqref{eq:lp-for-postcode} is feasible. Furthermore, the optimal value $v^*$ satisfies the bound
  \begin{align*}
    v^* \leq 4 \Delta^2.
  \end{align*}
\end{lemma}
\noindent See \Cref{subsubsec:proof-lemma-feasibility-post-coding} for the proof of this lemma. This lemma ensures feasibility of the linear program~\eqref{eq:lp-for-postcode} when the noise level is small enough, and the variance of the post-coding mapping $\correctionMap \circ \qtzadc \circ \channel$ is dominated by the quantization error $\Delta^2$.

\subsection{Scale-adaptive transformation}\label{subsec:scale-adaptive-transformation}
Another component in our algorithms is a scale-adaptive transformation $(\beta_\smallscale, \Psi_\smallscale)$, parametrized by a tuning parameter $\smallscale > 0$. Given an input scalar $x$, we define a pair of functions
\begin{subequations}
\begin{align}
  \beta_\smallscale (x) \mydefn \max \Big(0,  \lceil \log_2 \big( \smallscale^{-1} \abss{x} \big) \rceil \Big), \quad \mbox{and} \quad \Psi_\smallscale (x) = (1 - \Delta) x / \big( 2^{\beta_\smallscale (x)} \smallscale \big). \label{eq:defn-scale-adaptive}
\end{align}
In other words, we compare $|x|$ with a binary grid $(2^{k} \smallscale)_{k \geq 0}$, and sort it to the level corresponding to the index $\beta_\smallscale (x)$. We then re-scale the scalar with respect to this grid level to obtain $\Psi_\smallscale (x)$. Clearly, it is always guaranteed that $|\Psi_\smallscale (x)| \leq 1 - \Delta$, so that the output lies within the interval $[z_2, z_{q - 1}]$, applicable to the post-coding scheme described above. The tuning parameter $\smallscale$ is a small positive scalar chosen to address the trade-offs between communication complexity and statistical errors, which will be reflected in the theoretical guarantees.

For notational convenience, we also extend the mapping $\Psi$ to real vectors, in an entry-wise fashion. Concretely, given $x = \begin{bmatrix}
  x_1 & x_2& \cdots& x_\usedim
\end{bmatrix}^\top \in \real^\usedim$, we define
\begin{align}
  \Psi_\smallscale (x) \mydefn \begin{bmatrix}
    \Psi_\smallscale (x_1) & \cdots & \Psi_\smallscale (x_\usedim)
  \end{bmatrix}^\top, \quad \mbox{and} \quad   \beta_\smallscale (x) \mydefn \begin{bmatrix}
    \beta_\smallscale (x_1)  & \cdots & \beta_\smallscale (x_\usedim)
  \end{bmatrix}^\top. 
\end{align}
We also define the inverse operation that assembles the information transmitted through two channels into the original real-valued data:
\begin{align}
   A_\smallscale (\psi, b) \mydefn \frac{1}{1 - \Delta} 2^{b} \smallscale \cdot \psi,
\end{align}
and its vectorized version defined as entry-wise operations. This function assembles the information transmitted through two channels into the original real-valued data.
\end{subequations}

We need the following technical lemma about the over-the-air channel and the quantization process.
\begin{lemma}\label{lemma:joint-quantization-channel}
  Given a deterministic vector $u \in \real^\usedim$, define
  \begin{align*}
    \widehat{u} \mydefn A_\smallscale \Big( \correctionMap \circ \qtzadc \circ \channel \circ \qtzalg \big(  \Psi_\smallscale (u) \big), \beta_\smallscale (u) \Big),
  \end{align*}
  we have $\Exs [\widehat{u}] = u$, and the following inequalities hold
  \begin{align*}
    \Exs \big[ \vecnorm{\widehat{u} - u}{2}^2 \big] & \leq
    (4 v^* + \Delta^2) \cdot \big(4 \vecnorm{u}{2}^2 + \smallscale^2 \usedim \big).
  \end{align*}
\end{lemma}
\noindent See \Cref{subsubsec:proof-lemma-quantization-channel} for the proof of this lemma. This lemma ensures that the post-coded over-the-air channel $\correctionMap \circ \qtzadc \circ \channel \circ \qtzalg$ satisfies \Cref{eq:oracle-condition-intro}, thereby facilitating the convergence of stochastic optimization algorithms. It is worth noting that the variance of the post-coded over-the-air channel is adaptive to the transmitted signal $u$ itself.
\begin{figure}[htb]
  \centering
  \includegraphics[width=0.9\textwidth]{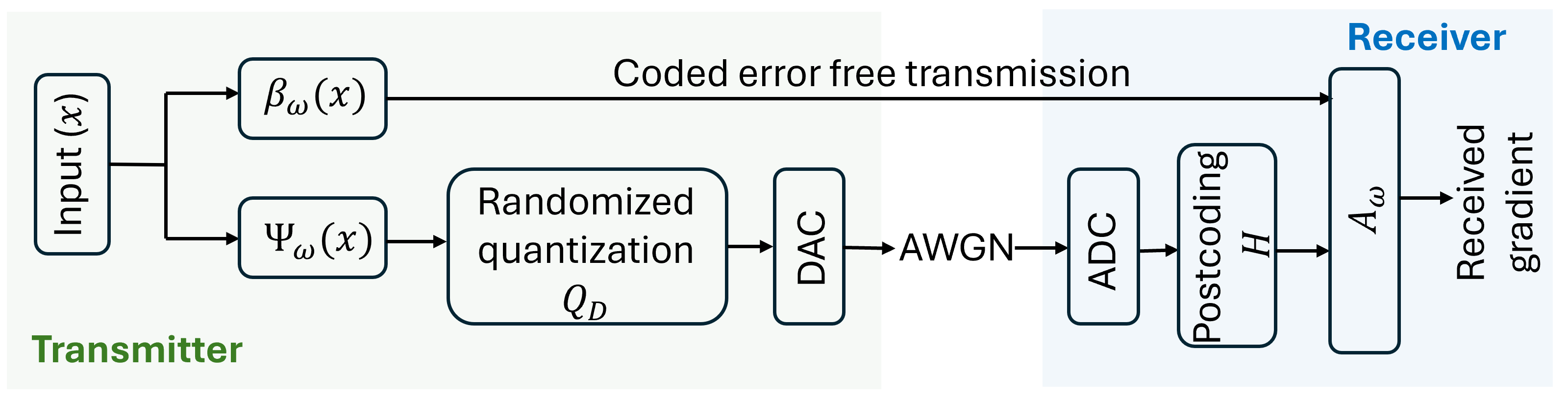}
  \caption{Block diagram of transmission process for physical channels}\label{fig:block-diagram-physical}
\end{figure}

In \Cref{fig:block-diagram-physical}, we describe the overall transmission process for physical channels, combining the quantization, channel noise, and post-coding steps, as well as the scale-adaptive transformation framework.

\subsection{Adaptive over-the-air SGD}

\begin{wrapfigure}{r}{0.4\linewidth}
  \centering
  \includegraphics[width=\linewidth]{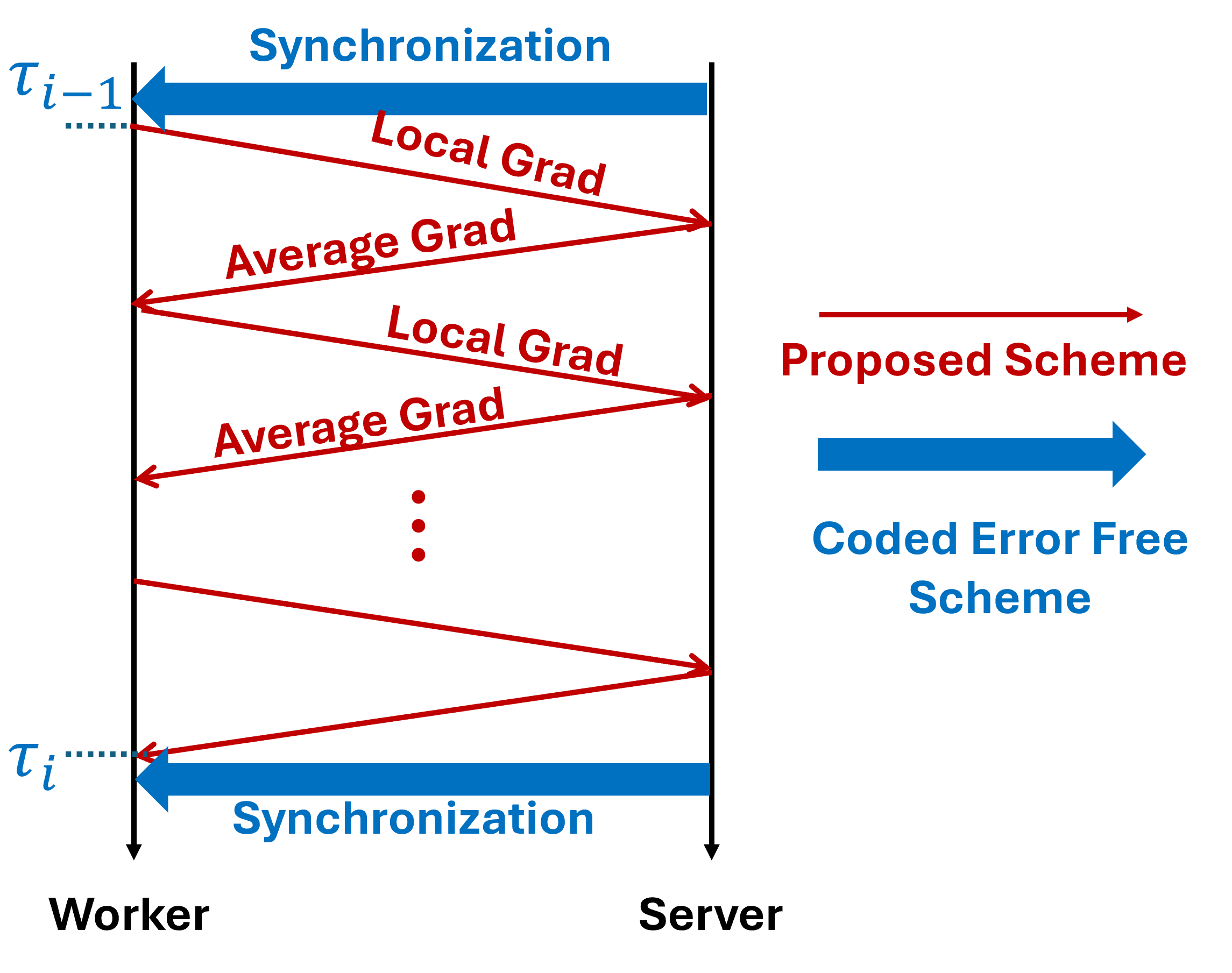}
  \caption{Federated learning algorithm overview}\label{fig:algorithm-overview}
\end{wrapfigure}

Given the data transmission routines established in the previous sections, we are now ready to describe the algorithmic framework for over-the-air federated learning. We work with a worker-server network architecture. In each round, the workers compute the local stochastic gradient and send it to the server, and the server broadcasts the aggregated gradient information. All the data transmissions in this process use the transmission scheme described in \Cref{fig:block-diagram-physical}, with the scale information transmitted through the coded channel, and the normalized values transmitted through the physical channel.

Additionally, we introduce a synchronization step to maintain stability. In particular, given an increasing sequence $\tau_1 < \tau_2 < \cdots < \tau_k < \cdots$ of time steps, the central machine broadcasts the current global model parameters $\theta_k$ to all workers at time step $\tau_k$, for each $k$. Upon receiving the synchronization message, the workers replace their local model parameters by the global ones. The synchronization steps do not need to be frequent. In our theoretical analysis, we will show a bound on the requirement for time intervals between synchronization steps.

In \Cref{fig:algorithm-overview}, we provide an illustration of the adaptive over-the-air SGD framework.
In \Cref{alg:server-base-version,alg:worker-base-version}, we present the detailed procedures for server and workers, respectively.

\begin{algorithm}[htb]
\caption{Adaptive over-the-air SGD: worker side}\label{alg:worker-base-version}
  \begin{algorithmic}
    \REQUIRE Initial point $\theta_0$, where $\theta_0^{(j)} = \theta_0$ for $j \in [m]$.
    \FOR{$k = 1,2, \cdots, \numobs$}
    \STATE Sample a local data $X_k^{(j)}$, and compute
    \begin{align*}
      \signalup_k^{(j)} \sim \qtzalg \big( \Psi_\smallscale \big(\nabla f (\theta_{k - 1}^{(j)}, X_k^{(j)}) \big) \big), \quad \mbox{and} \quad \beta_k^{(j)} \mydefn \beta_\smallscale \big(\nabla f (\theta_{k - 1}^{(j)}, X_k^{(j)}) \big),
    \end{align*}
    \STATE Transmit the real vector $\signalup_k^{(j)}$ through the over-the-air channel, and transmit the discrete vector $\beta_k^{(j)}$ to server through the coded channel.
    \STATE Receive a real vector $\widehat{\signaldown}_k^{(j)} \sim \correctionMap \circ \qtzadc \circ \channel (\signaldown_k)$ through post-coded over-the-air channel, and a discrete vector $\beta_k$ through coded channel; update local parameter
    \begin{align*}
      \theta_k^{(j)} = \theta_{k - 1}^{(j)} - \stepsize_k A_\smallscale \big( \widehat{\signaldown}_k^{(j)}, \beta_k \big).
    \end{align*}
    \IF{$k \in \{\tau_1, \tau_2, \cdots\}$}
    \STATE Receive $\theta_k$ from the server through the coded channel, and let $\theta_k^{(j)} = \theta_k$.
    \ENDIF
    \ENDFOR
  \end{algorithmic}
\end{algorithm}

\begin{algorithm}[htb]
  \caption{Adaptive over-the-air SGD: server side}\label{alg:server-base-version}
    \begin{algorithmic}
      \FOR{$k = 1,2, \cdots, \numobs$}
      \FOR{each machine $j  \in [m]$}
      \STATE Receive the transmitted data $\widehat{\signalup}_k^{(j)} \sim \correctionMap \circ \qtzadc \circ \channel (\signalup_k^{(j)})$ through post-coded over-the-air channel, and $\beta_k^{(j)}$ through coded channel.
      \ENDFOR
      \STATE Aggregate the received information and update local parameter
      \begin{align*}
       u_k = \frac{1}{m } \sum_{j = 1}^m A_\smallscale \big(\widehat{\signalup}_k^{(j)}, \beta_k^{(j)} \big), \quad \mbox{and} \quad \theta_k = \theta_{k - 1} - \stepsize u_k.
      \end{align*}
      \STATE Send $\signaldown_k = \qtzalg ( \Psi_\smallscale (u_k))$ through over-the-air channels, and $\beta_k = \beta_\smallscale (u_k)$ through coded channels.
      \ENDFOR
      \IF{$k \in \{\tau_1, \tau_2, \cdots\}$}
      \STATE Send $\theta_k$ to each workers through the coded channel.
      \ENDIF
    \end{algorithmic}
\end{algorithm}

\section{Theoretical guarantees}\label{sec:theory}
We present the theoretical guarantees for the adaptive over-the-air SGD algorithm. We will first present the results under strongly convex settings, and then move on to non-convex settings.
\subsection{Technical assumptions}\label{subsec:assumptions}
We make the following technical assumptions in our analysis.
\begin{subequations}
\begin{assumption}
  \label{assume:strong-convex-and-smooth}
  For any $\theta_1, \theta_2 \in \real^\usedim$, population-level loss function $F$ satisfies
  \begin{align}
   \mbox{$\smooth$-smoothness:} \qquad  F (\theta_1) - F (\theta_2) &\leq \inprod{\nabla F (\theta_1)}{\theta_1 - \theta_2} + \frac{\smooth}{2} \vecnorm{\theta_1 - \theta_2}{2}^2, \label{eq:smooth-assumption}\\
   \mbox{$\strongconvex$-strong convexity:} \qquad F (\theta_1) - F (\theta_2) &\geq \inprod{\nabla F (\theta_1)}{\theta_1 - \theta_2} + \frac{\strongconvex}{2} \vecnorm{\theta_1 - \theta_2}{2}^2.\label{eq:strongconvex-assumption}
  \end{align}
\end{assumption}
\end{subequations}
This assumption is standard in convex optimization literature. Note that we only require strong convexity and smoothness to hold for the population-level loss function $F$.
\begin{assumption}
  \label{assume:stochastic-lipschitz}
  The stochastic gradient oracle satisfies the moment bound
  \begin{align*}
   \Exs_{\Prob_j} \big[ \vecnorm{\nabla f (\theta, X)}{2}^{2} \big] \leq \sigstarj{j}^2 +  \sglip^2 \big( F (\theta) - F (\thetastar) \big), \quad \mbox{for any $\theta \in \real^\usedim$}.
  \end{align*}
  We also define the average noise level
  $\sigstar^2 \mydefn \tfrac{1}{m} \sum_{j = 1}^m \sigstarj{j}^2$.
\end{assumption}
This assumption is known as ``state-dependent noise'' condition in stochastic optimization literature~\cite{moulines2011non,ilandarideva2024accelerated}. It is more general than the standard bounded variance assumption, which requires $\Exs_{\Prob_j} [ \vecnorm{\nabla f (\theta, X)}{2}^{2}] \leq \sigstarj{j}^2$ for any $\theta$. The state-dependent noise condition is satisfied in many statistical learning problems, including generalized linear models. The noise level $\sigstar^2$ captures the average uncertainty in the gradient estimates across different workers, which governs the optimal statistical risk for machine learning problems.

\subsection{Results under strongly convex settings}
Under assumptions in \Cref{subsec:assumptions}, we have the following risk bounds for the adaptive over-the-air SGD algorithm, with last-iterate and average-iterate guarantees.

\begin{subequations}
To establish the theoretical results, we require the stepsize schedule $(\stepsize_k)_{k \geq 1}$ to satisfy
\begin{align}
  \stepsize_{k} \leq (1 + \stepsize_{k + 1} \strongconvex / 8) \stepsize_{k + 1}, \qquad \mbox{and} \qquad \stepsize_k \leq \frac{c_0}{\sglip^2 + \smoothness},
\end{align}
for some universal constant $c_0 > 0$.

Given the stepsize schedule, we need the synchronization times to satisfy the bounds for $i = 1,2, \cdots$
\begin{align}
  T (\tau_i) - T (\tau_{i - 1}) \leq \frac{1}{2 \smooth}, \quad \mbox{where } T (k) = \sum_{t = 1}^{k} \stepsize_t.\label{eq:sync-time-req}
\end{align}  
\end{subequations}
Under above setup, we can establish the theoretical results under strongly convex setup.
\begin{theorem}\label{thm:main-convex}
  Under Assumptions~\ref{assume:strong-convex-and-smooth} and~\ref{assume:stochastic-lipschitz}, given the stepsize sequence and synchronization times described above, for any $n \geq 1$, we have
  \begin{align*}
    \Exs \big[ \vecnorm{\theta_n - \thetastar}{2}^2 \big] \leq e^{- \frac{\strongconvex}{2} T (n)} \vecnorm{\theta_0 - \thetastar}{2}^2 + \frac{c \stepsize_n}{\strongconvex} \Big( \frac{\sigstar^2}{m} + (v^* + \Delta^2) \smallscale^2 \usedim \Big) .
  \end{align*}
\end{theorem}
\noindent See \Cref{subsec:proof-main-convex} for the proof of \Cref{thm:main-convex}.

A few remarks are in order. First, the bound in \Cref{thm:main-convex} is comparable to standard results for SGD in the centralized setting, where the convergence rate takes the form
\begin{align*}
  \Exs \big[ \vecnorm{\theta_n - \thetastar}{2}^2 \big] \leq e^{- \frac{\strongconvex}{2} T (n)} \vecnorm{\theta_0 - \thetastar}{2}^2 + c \stepsize_n \frac{\sigstar^2}{\strongconvex m} .
\end{align*}
Compared to this bound, the additional term $\frac{c \stepsize_n}{\strongconvex} (v^* + \Delta^2) \smallscale^2 \usedim $ in \Cref{thm:main-convex} accounts for the distributed nature of the optimization problem and the variability in the gradient estimates across different workers. This term can be made small by choosing a small $\smallscale$, which, however, increases the communication cost. This reflects the trade-off between communication cost and statistical accuracy in distributed optimization. Furthermore, the additional term also depends on the quantity $v^* + \Delta^2$, which reflects the impact of the SNR and hardware constraints.

By taking the stepsize choice $\stepsize_k \asymp \frac{1}{\sglip^2 + \smoothness + \strongconvex k}$, for $\numobs \gtrsim \frac{\sglip^2 + \smoothness}{\strongconvex}$, we have $T (n) \gtrsim \frac{1}{\strongconvex} \log n$, and first term in the bound is negligible. This leads to a sample complexity bound of
\begin{align*}
  \widetilde{O} \Big( \frac{\sglip^2 + \smoothness}{\strongconvex}  \Big) + \widetilde{O} \Big( \frac{1}{\strongconvex^2 \varepsilon^2} \Big\{ \frac{\sigstar^2}{m} + (v^* + \Delta^2) \smallscale^2 \usedim  \Big\} \Big)
\end{align*}
to achieve $ \Exs \big[ \vecnorm{\theta_n - \thetastar}{2}^2 \big] \leq \varepsilon^2$, where the $\widetilde{O} (\cdot)$ notation hides logarithmic factors. The first term is the optimization error, and the second term is the statistical error. The statistical error matches the minimax optimal rate $\frac{\sigstar^2}{\strongconvex m}$ when the term $(v^* + \Delta^2) \smallscale^2 \usedim$ is small enough.

Finally, we note that the synchronization requirement~\eqref{eq:sync-time-req} is not very stringent. For example, if we take the stepsize choice $\stepsize_k \asymp \frac{1}{\sglip^2 + \smoothness + \strongconvex k}$, then we have $T (k) \asymp \frac{1}{\strongconvex} \log k$, and the synchronization requirement~\eqref{eq:sync-time-req} is satisfied as long as $\tau_i / \tau_{i - 1} \leq c$ for some constant $c > 1$. In other words, the synchronization times can be chosen to be geometrically increasing. On the other hand, if we choose a fixed stepsize $\stepsize_k = \stepsize > 0$, then the synchronization requirement~\eqref{eq:sync-time-req} requires $\tau_i - \tau_{i - 1} \leq \frac{1}{2 \smooth \stepsize}$, i.e., the synchronization steps need to be performed at a constant frequency that is inverse proportional to the stepsize.

\subsection{Results under non-convex settings}
Similarly, we can also establish results for finding stationary points of a non-convex function.

\begin{assumption}
  \label{assume:non-convex-sglip}
  The stochastic gradient oracle satisfies the moment bound
  \begin{align*}
   \Exs_{\Prob_j} \big[ \vecnorm{\nabla f (\theta, X)}{2}^{2} \big] \leq \sigstarj{j}^2 + \lambda \vecnorm{\nabla F (\theta)}{2}^2, \quad \mbox{for any $\theta \in \real^\usedim$},
  \end{align*}
\end{assumption}
This assumption is an extension of the state-dependent noise variance assumption~\ref{assume:stochastic-lipschitz} to the non-convex setting, which, once again, allows the stochastic gradient variance to be state-dependent and unbounded.

Aligned with standard practice in non-convex optimization literature, we measure the quality of a solution $\theta$ by the squared gradient norm $\vecnorm{\nabla F (\theta)}{2}^2$. To present the result, we need to introduce a random variable $R$, which takes values in $\{0,1,\cdots, n - 1\}$, with probabilities proportional to the stepsizes,
\begin{align*}
  \Prob ( R = k ) = \frac{\stepsize_{k + 1}}{\sum_{t = 1}^{n } \stepsize_t}.
\end{align*}
Now we can state the main result for the non-convex setting.
\begin{theorem}\label{thm:non-convex}
  Under \Cref{eq:smooth-assumption} and~\Cref{assume:non-convex-sglip}, when the stepsizes satisfy $\stepsize_k \leq \tfrac{c_0}{\sglip^2 + \smoothness}$, and the synchronization times satisfy \Cref{eq:sync-time-req}, we have
  \begin{align*}
    \Exs \big[ \vecnorm{\nabla F (\theta_R)}{2}^2 \big] \leq \frac{F (\theta_0) - F_{\min} +  c\smoothness \Big\{  \frac{\sigstar^2}{m} +  (v^* + \Delta^2) \smallscale^2 \usedim \Big\} \cdot \sum_{k = 1}^n \stepsize_k^2}{\sum_{k = 1}^n \stepsize_k},
  \end{align*}
  where $c_0, c > 0$ are universal constants, and $F_{\min} = \inf_{\theta \in \real^\usedim} F (\theta)$.
\end{theorem}
\noindent See \Cref{subsec:proof-thm-nonconvex} for the proof of this theorem. A few remarks are in order. First, similar to \Cref{thm:main-convex}, the bound in \Cref{thm:non-convex} is comparable to standard results for SGD in the centralized setting, with an additional term $\smoothness (v^* + \Delta^2) \smallscale^2 \usedim $ accounting for communication channels and quantization hardwares. By taking the stepsize $\stepsize_k \asymp 1 / \sqrt{\numobs}$, we can achieve an $O(\varepsilon^{-4})$ sample complexity bound to achieve $\Exs [ \vecnorm{\nabla F (\theta_R)}{2}^2 ] < \varepsilon^2$, which is standard in the analysis of SGD in non-convex settings. Note that the synchronization requirement~\eqref{eq:sync-time-req} is the same as that in \Cref{thm:main-convex}. Under the $1 / \sqrt{\numobs}$ stepsize choice, this requirement becomes $\tau_i - \tau_{i - 1} \asymp \sqrt{\numobs}$. In other words, we only need $O(\sqrt{\numobs})$ broadcasting steps in the total $\numobs$ iterations.

\section{Simulation studies}\label{sec:simulation}
In this section, we conduct simulation studies to validate the theoretical findings of our paper. We consider a simple federated learning problem of image classification on the MNIST dataset. The training data is distributed across $m = 10$ worker machines, and each worker has the data for each digit class. The neural architecture we use is a 4-layer convolutional neural network (CNN), with two convolutional layers and two fully connected layers. The total number of parameters is $\usedim = 1625866$. We use the cross-entropy loss function for training, with stepsize $\stepsize = 0.01$, and the batch size is set to $64$ for each worker.

We compare 5 different transmission schemes in our experiments:
\begin{itemize}
  \item \textbf{Coded}: In this scheme, all the information is transmitted through the coded channel, using 32-bit floating point precision.
  \item \textbf{Noisy}: In this scheme, the information is transmitted directly through the over-the-air channel described in \Cref{subsec:physical-constraints}, which includes the DAC unit, the AWGN channel, and the ADC unit.
  \item \textbf{Postcode:} In this scheme, we apply the post-coding and scale-adaptive transformation techniques in \Cref{subsec:post-coding} and \Cref{subsec:scale-adaptive-transformation}, to transmit each parameter in an unbiased manner. A simple distributed SGD algorithm is used for training, without the synchronization step described in \Cref{alg:server-base-version,alg:worker-base-version}.
  \item \textbf{Sync:} In this scheme, we run the distributed optimization framework in \Cref{alg:server-base-version,alg:worker-base-version}, with the global model parameters being synchronized across all workers at communication rounds $\tau_1, \tau_2, \cdots$. The transmission is performed over the noisy channel described in \Cref{subsec:physical-constraints}, without the post-coding and scale-adaptive transformation techniques. In our simulation studies, the syncrhonization time interval is chosen to be $100$, i.e., $\tau_k = 100 k$ for $k = 1,2, \cdots$.
  \item \textbf{Ours:} This scheme corresponds to the full algorithms described in \Cref{alg:server-base-version,alg:worker-base-version}, incorporating post-coding, scale-adaptive transformation, and synchronization.
\end{itemize}

\begin{figure}[htb]
  \centering
  \begin{minipage}{0.48\textwidth}
    \centering
    \includegraphics[width=\textwidth]{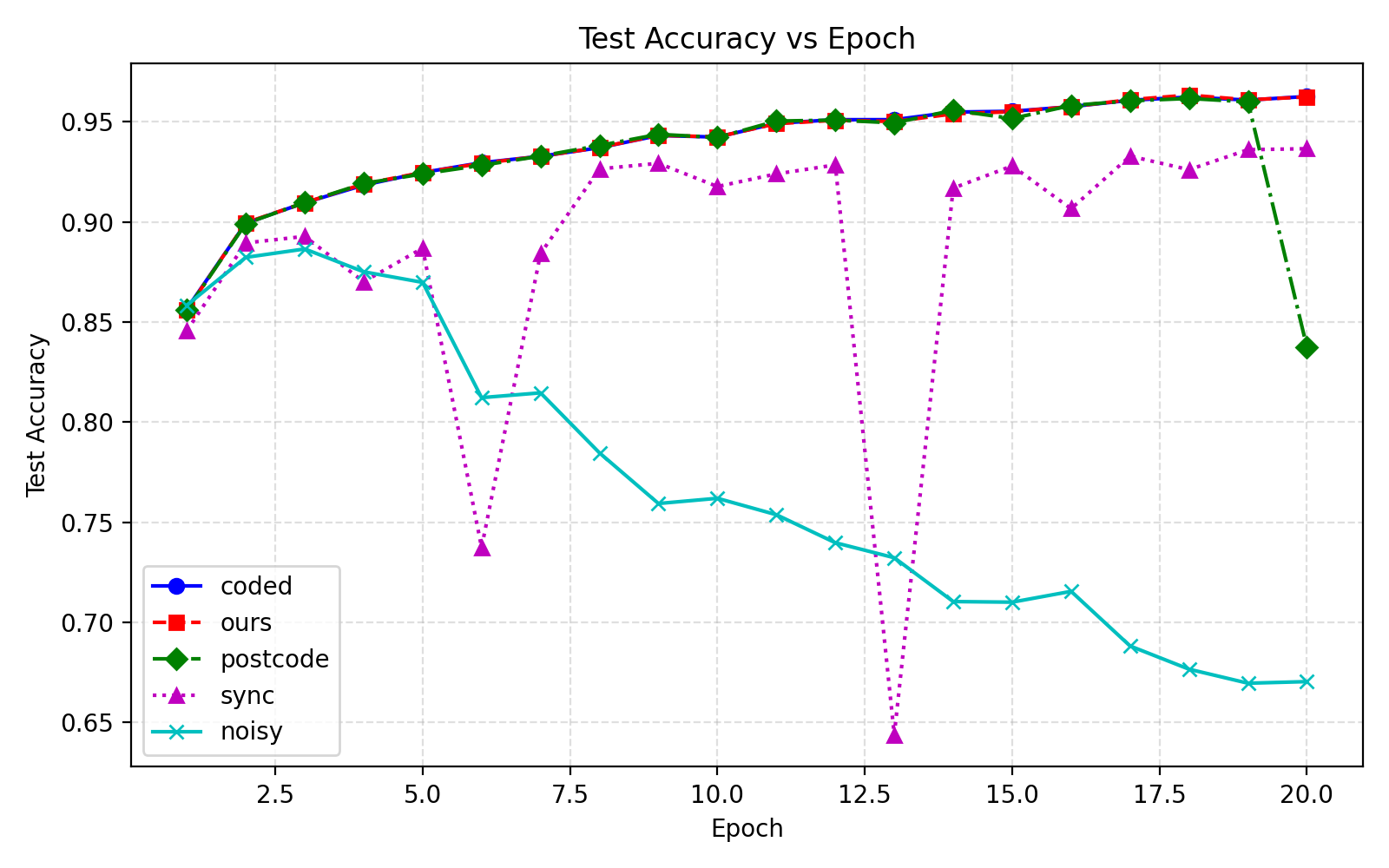}
    \subcaption*{(a) Test accuracy vs. epoch (High SNR)}
  \end{minipage}
  \hfill
  \begin{minipage}{0.48\textwidth}
    \centering
    \includegraphics[width=\textwidth]{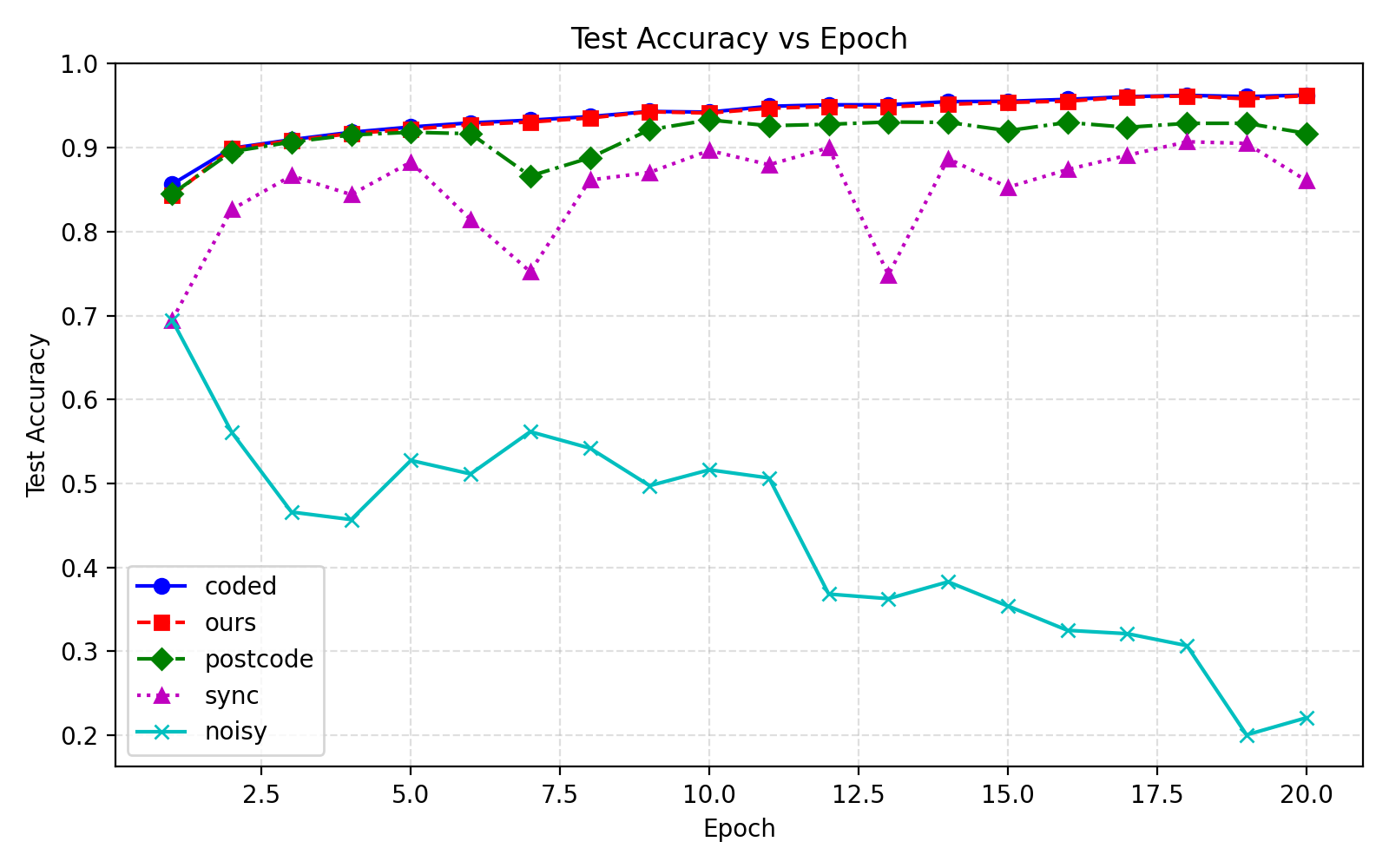}
    \subcaption*{(b) Test accuracy vs. epoch (Low SNR)}
  \end{minipage}
  \vspace{0.5em}
  \begin{minipage}{0.48\textwidth}
    \centering
    \includegraphics[width=\textwidth]{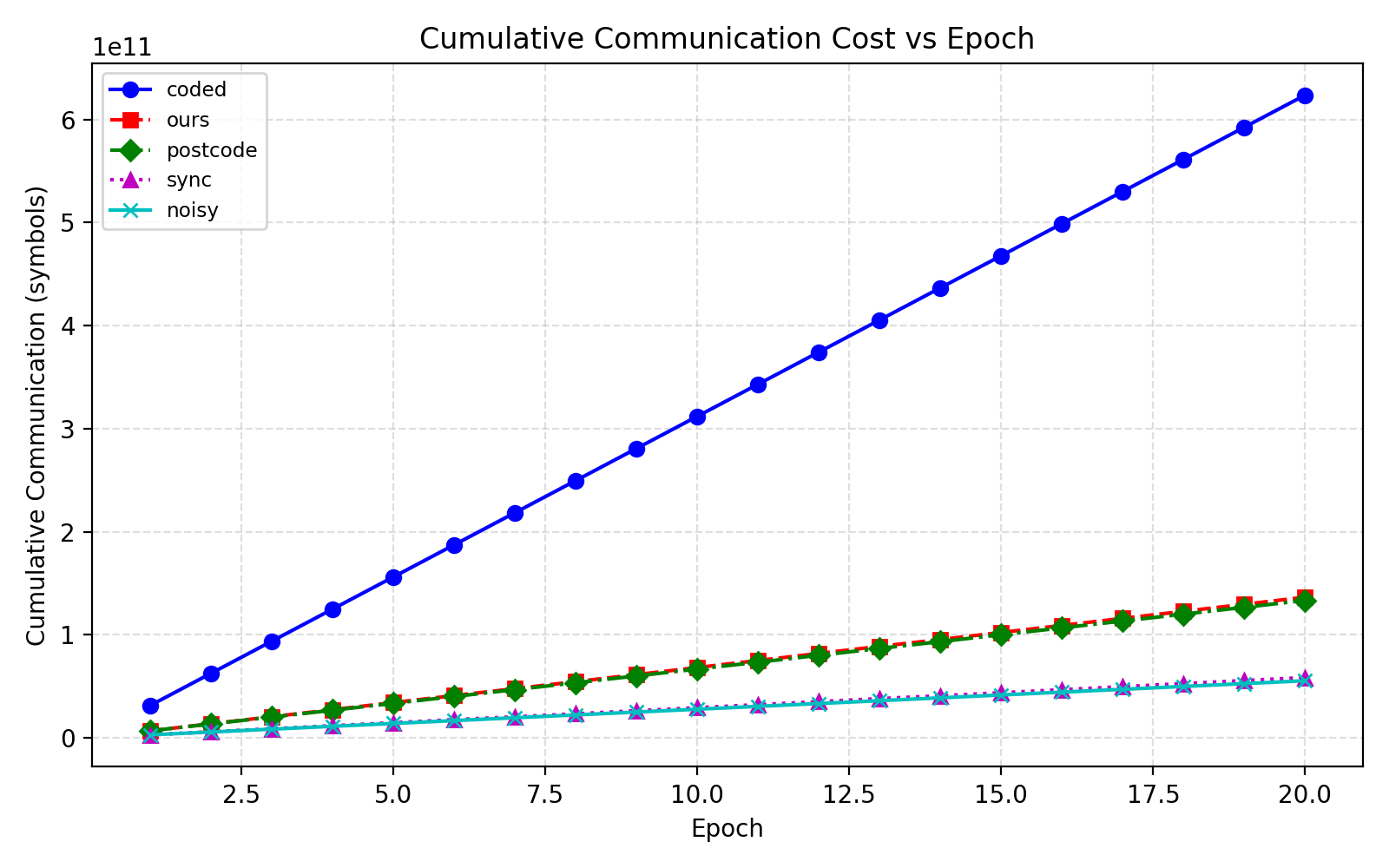}
    \subcaption*{(c) Communication cost vs. epoch (High SNR)}
  \end{minipage}
  \hfill
  \begin{minipage}{0.48\textwidth}
    \centering
    \includegraphics[width=\textwidth]{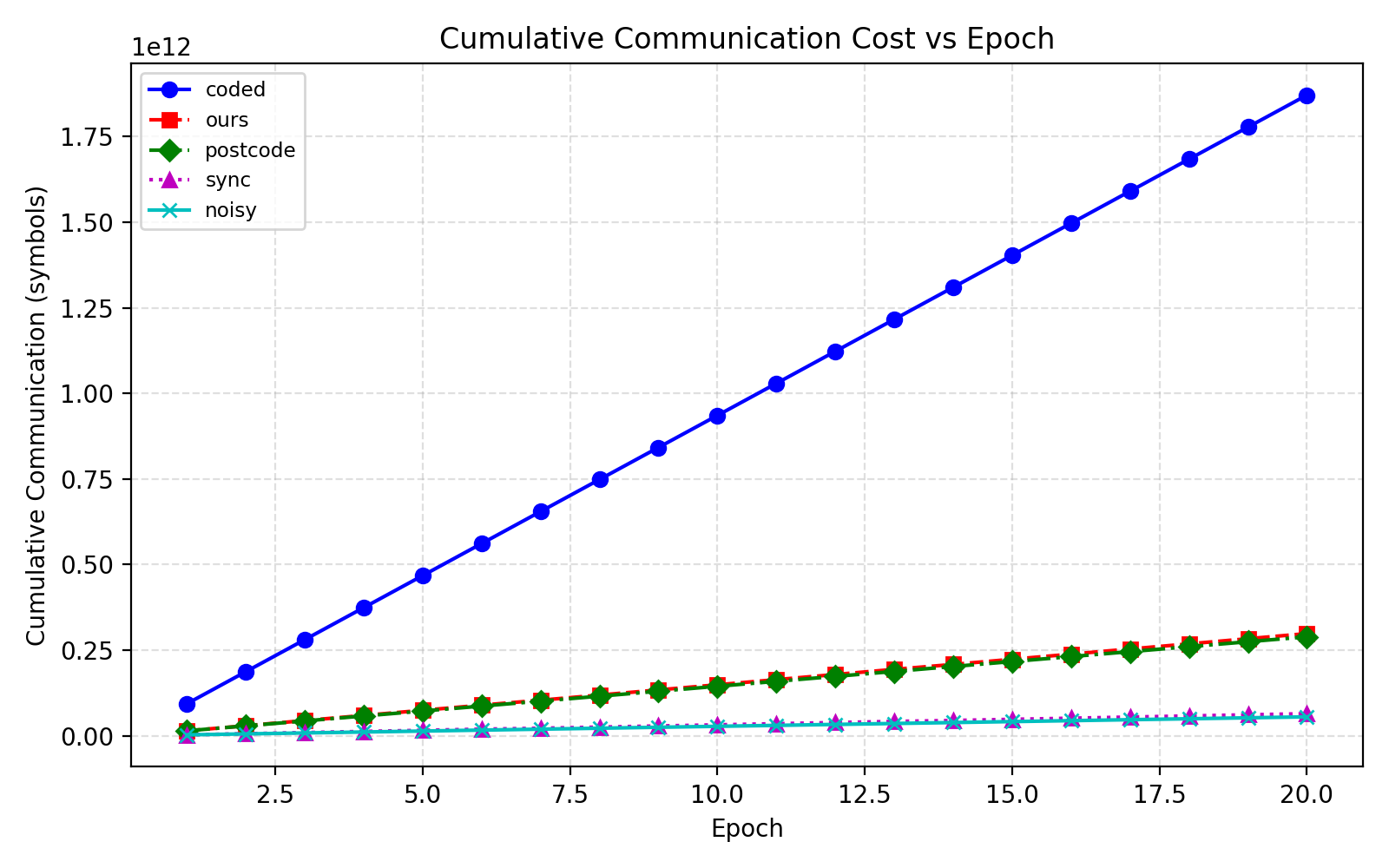}
    \subcaption*{(d) Communication cost vs. epoch (Low SNR)}
  \end{minipage}
  \caption{Simulation results for federated learning over physical channels. (a) and (b): Test accuracy over epochs for high and low SNR regimes, respectively. (c) and (d): Communication cost over epochs for high and low SNR regimes, respectively.}
  \label{fig:simulation-results}
\end{figure}
We test these 5 different methods under simulated communication channels. To ensure a fair comparison, we require the average  signal power to be the same for coded and noisy channels. We consider two different SNR regimes in our experiments:
\begin{itemize}
  \item High SNR regime: we let $\sigma_c = 0.05$ in the communication channel, and the quantization levels are set to $q = 16$. For coded communication, we consider a PAM-8 modulation with Gray mapping.\footnote{In many communication systems, QAM modulations are used. In such a case, we use the real and imaginary parts to encode two different PAM symbols, so the number of symbols is halved for both coded and over-the-air channels.} In this case, the SNR is approximately 19.5dB, and the pre-FEC BER is about $1.04 \times 10^{-3}$ (see e.g.~\cite{proakis2001digital}). Following industry standards for FEC overhead~\cite{agrell2018information,ieee802.3-2018}, we assume an FEC overhead of $5.8\%$.
  \item Low SNR regime: we let $\sigma_c = 0.2$ in the communication channel, and the quantization levels are set to $q = 8$. In this case, the SNR is approximately 5.5dB. For coded communication, we consider a BPSK modulation, leading to a pre-FEC BER of $3.86 \times 10^{-3}$ (see e.g.~\cite{proakis2001digital}). Following industry standards for FEC overhead~\cite{agrell2018information,ieee802.3-2018}, we assume an FEC overhead of $5.8\%$.
\end{itemize}
Now we are ready to present our simulation results. We run the algorithms for a total of 20 epochs, and we report the test accuracy and communication overhead for each transmission scheme. In \Cref{fig:simulation-results}, we present the results for both high and low SNR regimes. The communication cost is measured in terms of the total number of symbols transmitted through the channel, which include both coded and physical transmissions.

From \Cref{fig:simulation-results} (a) and (b), we observe that the test accuracy of our method matches the performance of coded transmission schemes in both high and low SNR regimes. In contrast, if we use the noisy physical channel directly, or use the post-coding approach or the synchronization framework alone, the test accuracy becomes unstable and drops significantly. Indeed, in our simulation, the test accuracy of our method and the coded transmission results differs by only $0.07\%$. On the other hand, the communication cost of our method is consistently lower than that of the coded transmission schemes, leading to more than 5 times savings in the number of symbols transmitted, as shown in \Cref{fig:simulation-results} (c) and (d). The post-coding and scale-adaptive transformation leads to some overhead in communication cost, as observed in \Cref{fig:simulation-results} (c) and (d), but this is outweighed by the gains in test accuracy.

\section{Proofs}\label{sec:proofs}
We present the proofs of the main results in this section.
\subsection{Proofs about the transmission channels}
In this section, we collect the proofs of the technical lemmas about the transmission channels, the quantization process, and the post-coding procedure.
\subsubsection{Proof of \Cref{lemma:feasibility-post-coding}}\label{subsubsec:proof-lemma-feasibility-post-coding}
We show feasibility and boundedness of the linear program~\eqref{eq:lp-for-postcode} by direct construction. Given a vector $\zeta \in \real^{q - 2}$, we define the matrix $H (\zeta)$ as follows (for simplicity, we index the elements of $\zeta$ from $2$ to $q - 1$).
\begin{align*}
  H (\zeta) \mydefn \begin{bmatrix}
    1 & 0 & 0 & 0 & \cdots & 0\\
    \frac{1 - \zeta_2}{3} & \frac{1}{3} & \frac{1 + \zeta_2}{3} & 0 & \cdots & 0\\
    0 & \frac{1 - \zeta_3}{3} & \frac{1}{3} & \frac{1 + \zeta_3}{3} & \cdots & 0\\
    \vdots & \vdots & \vdots & \ddots & \vdots\\
    0 & \cdots &0& \frac{1 - \zeta_{q - 1}}{3} & \frac{1}{3} & \frac{1 + \zeta_{q - 1}}{3} \\
    0 & 0 &0 & 0 & \cdots & 1
  \end{bmatrix} \in \real^{q \times q}.
\end{align*}
Clearly, the matrix $H (\zeta)$ satisfies the constraint~\eqref{eq:lp-for-postcode-constraint-1} whenever $\vecnorm{\zeta}{\infty} \leq 1$. Now we study the constraint~\eqref{eq:lp-for-postcode-constraint-2}. First, since the quantization levels $(z_1, z_2, \cdots, z_q)$ are equi-spaced, for each $j \in \{2,3, \cdots, q - 2\}$, we note that
\begin{align*}
  \coordinate_j^\top P H (0) z = P_{j, 1} z_1 + P_{j, q} z_q + \sum_{i = 2}^{q - 1} P_{j,i} \cdot \frac{1}{3} (z_i + z_{i - 1} + z_{i + 1}) = \Exs \big[ \qtzadc \circ \channel (z_j) \big],
\end{align*}
and consequently
\begin{align*}
  \coordinate_j^\top P H (\zeta) z - z_j &= \coordinate_j^\top P \big( H (\zeta) - H (0) \big) z + \Exs \big[ \qtzadc \circ \channel (z_j) \big] - z_j\\
  &= \sum_{i = 2}^{q - 1} P_{j, i} \frac{1}{3} \zeta_i (z_{i + 1} - z_{i - 1}) + \Exs \big[ \qtzadc \circ \channel (z_j) \big] - z_j\\
  &= \frac{2 \Delta}{3} \cdot \sum_{i = 2}^{q - 1} P_{j, i} \zeta_i + \Exs \big[ \qtzadc \circ \channel (z_j) \big] - z_j.
\end{align*}
Define the matrix $P^* \in \real^{(q - 2) \times (q - 2)}$ as the restriction of the matrix $P$ to the rows and columns corresponding to the indices $\{2,3, \cdots, q - 1\}$. We claim that the matrix $P^*$ is invertible, satisfying the bound
\begin{align}
  \matsnorm{(P^*)^{-1}}{\infty \rightarrow \infty} \leq 3, \quad \mbox{when } \sigma_c \leq \frac{\Delta}{2}.\label{eq:Pstar-inverse-bound-in-lemma-lp-feasibility-proof}
\end{align}
We prove this result at the end of this section. Taking this operator norm bound as given, we define the vector $\zeta^*$ as
\begin{align*}
  \zeta^* = \frac{3}{2 \Delta} \cdot (P^*)^{-1} \cdot \big(z_j - \Exs \big[ \qtzadc \circ \channel (z_j) \big] \big)_{j = 2}^{q - 1}.
\end{align*}
To establish feasibility, we need to show that $\vecnorm{\zeta^*}{\infty} \leq 1$. By \Cref{eq:Pstar-inverse-bound-in-lemma-lp-feasibility-proof}, we have
\begin{align*}
  \vecnorm{\zeta^*}{\infty} \leq \frac{3}{2 \Delta} \cdot \matsnorm{(P^*)^{-1}}{\infty \rightarrow \infty} \cdot \max_{j = 2,3, \cdots, q - 1} \big| z_j - \Exs \big[ \qtzadc \circ \channel (z_j) \big] \big| \leq \frac{9}{2 \Delta} \max_{j = 2,3, \cdots, q - 1} \big| z_j - \Exs \big[ \qtzadc \circ \channel (z_j) \big] \big|.
\end{align*}
For the channel and ADC unit, we note that for any fixed $y \in [-1, 1]$, by symmetry of the normal density, we have
\begin{align*}
  \abss{\Exs [\qtzadc \circ \channel (y)] - y} &= \abss{\int_{-\infty}^\infty \big( \qtzadc (y + z) - y \big) \cdot \frac{1}{\sqrt{2\pi \sigma_c^2}} e^{\frac{- z^2}{2 \sigma_c^2}} dz} \\
  &= \abss{\int_0^\infty \big( \qtzadc (y + z) + \qtzadc (y - z) - 2 y \big) \cdot \frac{1}{\sqrt{2\pi \sigma_c^2}} e^{\frac{- z^2}{2 \sigma_c^2}} dz}\\
  &\leq 2 \int_{1 - |y|}^{+ \infty}  \frac{z}{\sqrt{2\pi \sigma_c^2}} e^{\frac{- z^2}{2 \sigma_c^2}} dz =\sqrt{\frac{2 \sigma_c^2}{\pi}} \exp \Big( \frac{- (1 - |y|)^2}{2 \sigma_c^2} \Big).
\end{align*}
When $\sigma_c \leq \Delta / 2$, this implies that
\begin{align*}
  \max_{j = 2,3, \cdots, q - 1} \big| z_j - \Exs \big[ \qtzadc \circ \channel (z_j) \big] \big| \leq \sigma_c \sqrt{\frac{2 }{\pi}} \exp \Big( \frac{- \Delta^2}{2 \sigma_c^2} \Big) \leq e^{-2} \sqrt{\frac{2}{\pi}} \sigma_c \leq \frac{1}{9} \Delta.
\end{align*}
Consequently, we have $\vecnorm{\zeta^*}{\infty} \leq 1 / 2 < 1$. So the matrix $H (\zeta^*)$ satisfies the constraints~\eqref{eq:lp-for-postcode-constraint-1} and \eqref{eq:lp-for-postcode-constraint-2} simultaneously, and the linear program~\eqref{eq:lp-for-postcode} is feasible. To derive the optimal value bound, we note that the variance under $H (\zeta^*)$ satisfies
\begin{align*}
  \var \big( \correctionMap \circ \qtzadc \circ \channel (z_j) \big) &\leq 2 \Exs \big[ \abss{\correctionMap \circ \qtzadc \circ \channel (z_j) - \qtzadc \circ \channel (z_j)}^2 \big] + 2\Exs \big[ \abss{\qtzadc \circ \channel (z_j) - z_j}^2 \big]\\
  &\leq 2 \Delta^2 +   2 \Exs \big[ \abss{\qtzadc \circ \channel (z_j) - z_j}^2 \big],
\end{align*}
where we use Young's inequality and almost-sure boundedness of the random mapping $\correctionMap$. 
To bound the second term, we define an auxiliary function $\qtzadc' (x) \mydefn \arg\min_{i \Delta : i \in \mathbb{Z}} \abss{x - i \Delta}$. For any $x \in [-1,1]$, we clearly have
\begin{align*}
  \Exs \big[\abss{\qtzadc \circ \channel (x) - x}^2 \big]  \leq \Exs \big[\abss{\qtzadc' \circ \channel (x) - x}^2 \big].
\end{align*}
and we have
\begin{align*}
  \Exs \big[\abss{\qtzadc' \circ \channel  (x) - x}^2 \big] &\leq 2 \Exs \big[ \abss{\qtzadc' \circ \channel  (x) -  \channel  (x)}^2 \big] + 2 \Exs \big[ \abss{ \channel  (x) - x}^2 \big]\\
  &\leq 2 \cdot \frac{\Delta^2}{4} + 2 \sigma_c^2.
\end{align*}
Putting them together, we noting that $\sigma_c \leq \Delta / 2$, we have
\begin{align*}
  \var \big( \correctionMap \circ \qtzadc \circ \channel (z_j) \big) \leq 4 \Delta^2,
\end{align*}
which completes the proof of \Cref{lemma:feasibility-post-coding}.

\paragraph{Proof of \Cref{eq:Pstar-inverse-bound-in-lemma-lp-feasibility-proof}:} For notational consistency, we index the elements of $(q- 2)$-dimensional vectors from $2$ to $q - 1$. For any vector pair $x, y \in \real^{q - 2}$ satisfying $x = P^* y$, we note that
\begin{align*}
  |x_i| = \abss{\sum_{j = 2}^{q - 1} P_{i,j} y_j} \geq P_{i, i} |y_i| - \sum_{j \neq i} P_{i, j} |y_j| \geq  P_{i, i} |y_i| - \vecnorm{y}{\infty} \sum_{j \neq i} P_{i, j}.
\end{align*}
Taking $i_0$ to be the index with the largest $|y_i|$, we have
\begin{align*}
  \vecnorm{x}{\infty} \geq |x_{i_0}| \geq P_{i_0, i_0} |y_{i_0}| - \vecnorm{y}{\infty} \sum_{j \neq i_0} P_{i_0, j} \geq \vecnorm{y}{\infty} \cdot \Big\{ P_{i_0, i_0} -  \sum_{j \neq i_0} P_{i_0, j} \Big\} = \vecnorm{y}{\infty} \big(2P_{i_0, i_0} - 1 \big).
\end{align*}
By definition, we have
\begin{align*}
  P_{i_0, i_0} = \Phi \big( \frac{\Delta}{2 \sigma_c} \big) -  \Phi \big( - \frac{\Delta}{2 \sigma_c} \big) \geq \frac{2}{3}, \quad \mbox{when } \sigma_c \leq \frac{\Delta}{2}.
\end{align*}
Under this condition, we have that $\vecnorm{P^* y}{\infty} \geq \frac{1}{3} \vecnorm{y}{\infty}$, for any $y \in \real^{q - 2}$. This implies that the matrix $P^*$ is invertible, and the operator norm $\matsnorm{(P^*)^{-1}}{\infty \rightarrow \infty} \leq 3$.

\subsubsection{Proof of \Cref{lemma:joint-quantization-channel}}\label{subsubsec:proof-lemma-quantization-channel}
We first prove unbiasedness. For each scalar $x$ satisfying $|x| \leq 1 - \Delta$, suppose that $x \in [z_i, z_{i + 1})$ for some $i \in \{2,3, \cdots, q - 2\}$. By definition, we have
\begin{align*}
  \Exs [\qtzalg (x)] = z_i \cdot \Prob \big( \qtzalg (x) = z_i \big) + z_{i + 1} \cdot \Prob \big( \qtzalg (x) = z_{i + 1} \big) = x.
\end{align*}
Furthermore, note that $\qtzalg (x) \in \{z_i, z_{i + 1}\}$ for $i \in \{2,3, \cdots, q - 2\}$. So we have $\qtzalg (x) \in \{z_2, \cdots, z_{q - 1}\}$ almost surely. By the linear program construction~\eqref{eq:lp-for-postcode}, we have
\begin{align*}
  \Exs \big[\correctionMap \circ \qtzadc \circ \channel \circ \qtzalg (x) \mid \qtzalg (x) \big] & = \qtzadc (x),
\end{align*}
and consequently, $\Exs [\correctionMap \circ \qtzadc \circ \channel \circ \qtzalg (x)] = x$. By construction, each coordinate of $\Psi_\smallscale (u)$ is bounded by $1 - \Delta$. Applying the above argument to each coordinate, we have $\Exs [\widehat{u}] = u$.

Now we turn to bound the variance. For each coordinate $i \in [\usedim]$, we note that
\begin{align}
  \Exs \big[ \abss{ \widehat{u}_i - u_i}^2 \big] &= 2^{2 \beta_\smallscale (u_i)} \frac{\smallscale^2}{(1 - \Delta)^2} \Exs \Big[  \abss{ \correctionMap \circ  \qtzadc \circ \channel \circ \qtzalg \big( \Psi_\smallscale (u_i) \big) -  \Psi_\smallscale (u_i) }^2  \Big] \nonumber\\
  &\leq \frac{2^{2 \beta_\smallscale (u_i)} \smallscale^2}{(1 - \Delta)^2} \Big\{ \Exs \big[ \abss{ \correctionMap \circ  \qtzadc \circ \channel \circ \qtzalg \big( \Psi_\smallscale (u_i) \big) -  \qtzalg \big(\Psi_\smallscale (u_i) \big)}^2 \big] + \var \big( \qtzalg (\Psi_\smallscale (u_i)) \big) \Big\}.\label{eq:variance-decomposition-in-lemma-quantization-channel}
\end{align}
By construction, we have that
\begin{align*}
  2^{\beta_\smallscale (u_i)} \smallscale \leq \max \big( 2 |u_i|, \smallscale \big).
\end{align*}
For the variance terms, we note that
\begin{align*}
 & \var \big( \qtzalg (\Psi_\smallscale (u_i)) \big) \leq \sup_{|x| \leq 1 - \Delta} \var (\qtzalg (x)) \leq \frac{\Delta^2}{4}.\\ 
  &\Exs \big[ \abss{ \correctionMap \circ  \qtzadc \circ \channel \circ \qtzalg \big( \Psi_\smallscale (u_i) \big) -  \qtzalg \big(\Psi_\smallscale (u_i) \big)}^2 \big] \leq \max_{i \in \{2,\cdots, q - 2\}} \Exs \big[ \abss{ \correctionMap \circ  \qtzadc \circ \channel (z_i) -  z_i}^2 \big] \leq v^*.
\end{align*}
Substituting these bounds into \Cref{eq:variance-decomposition-in-lemma-quantization-channel}, we have
\begin{align*}
  \Exs \big[ \abss{ \widehat{u}_i - u_i}^2 \big] &\leq 2^{2 \beta_\smallscale (u_i)} \frac{\smallscale^2}{(1 - \Delta)^2} \Big\{ v^* + \frac{\Delta^2}{4} \Big\}\\
  &\leq \big(4 v^* + \Delta^2 \big) \cdot \big(4 |u_i|^2 + \smallscale^2 \big).
\end{align*}
Aggregating the bounds for all the $\usedim$ coordinates, we conclude that
\begin{align*}
  \Exs \big[ \vecnorm{ \widehat{u} - u}{2}^2 \big] \leq (4 v^* + \Delta^2) \cdot \big( \smallscale^2 \usedim + 4 \vecnorm{u}{2}^2 \big).
\end{align*}

\subsection{Proof of \Cref{thm:main-convex}}\label{subsec:proof-main-convex}
We start with the one-step error decomposition
\begin{align}
  \Exs \big[ \vecnorm{\theta_k - \thetastar}{2}^2 \big] = \Exs \big[ \vecnorm{\theta_{k - 1} - \thetastar}{2}^2 \big] - 2 \stepsize_k \Exs \big[ \inprod{\theta_{k - 1} - \thetastar}{u_k} \big] + \stepsize_k^2 \Exs \big[\vecnorm{u_k}{2}^2 \big].\label{eq:one-step-error-decomposition-in-thm-main-convex}
\end{align}
We define the average disagreement between the local models and the global model as
\begin{align*}
  D_k \mydefn \frac{1}{m} \sum_{j=1}^m \Exs \big[ \vecnorm{\theta_{k - 1}^{(j)} - \theta_{k - 1}}{2}^2 \big].
\end{align*}
 The proof crucially relies on bounds on the bias and variance of the aggregated stochastic gradient $u_k$, which are summarized in the following lemmas.
 \begin{lemma}\label{lemma:main-recursion-cross-term-bound}
  Under the setup of \Cref{thm:main-convex}, we have
  \begin{multline*}
    \Exs \big[ \inprod{\theta_{k - 1} - \thetastar}{u_k} \big] \geq \frac{\strongconvex}{4} \Exs \big[ \vecnorm{\theta_{k - 1} - \thetastar}{2}^2 \big] + \frac{1}{2 m} \sum_{j = 1}^m \Exs \big[ F (\theta_{k - 1}^{(j)}) - F (\thetastar) \big]\\
    + \frac{1}{8 \smoothness m} \sum_{j = 1}^m \Exs \big[ \vecnorm{\nabla F (\theta_{k - 1}^{(j)})}{2}^2 \big]
   - 3 \smoothness D_{k - 1}.
  \end{multline*}
 \end{lemma}
 \noindent See \Cref{subsubsec:proof-lemma-main-recursion-cross-term-bound} for the proof of this lemma.

 \begin{lemma}\label{lemma:main-recursion-variance-bound}
  Under the setup of \Cref{thm:main-convex}, we have
  \begin{multline*}
    \Exs [\vecnorm{u_k}{2}^2]  \leq c \frac{\sigstar^2}{m} + c \frac{v^* + \Delta^2}{m} \smallscale^2 \usedim + \frac{c}{m} \sum_{j = 1}^m \Exs \big[ \vecnorm{\nabla F (\theta_{k - 1}^{(j)})}{2}^2 \big] + c \frac{\sglip^2}{m^2} \sum_{j = 1}^{m} \Exs \big[ F (\theta_{k - 1}^{(j)}) - F (\thetastar) \big].
  \end{multline*}
  where $c$ is a universal constant.
 \end{lemma}
  \noindent See \Cref{subsubsec:proof-lemma-main-recursion-variance-bound} for the proof of this lemma.

\begin{lemma}\label{lemma:worker-async-bound}
  Under the setup of \Cref{thm:main-convex}, for $k \in [\tau_{i - 1}, \tau_i )$, we have
  \begin{align*}
    D_k \leq   c_1 ( v^* + \Delta^2)  \sum_{t = \tau_{i - 1} + 1}^{k} \stepsize_t^2 \Big\{ \frac{\sigstar^2}{m} + \smallscale^2 \usedim +  \frac{1}{m} \sum_{j = 1}^m \Exs \big[ \vecnorm{\nabla F (\theta_{t - 1}^{(j)})}{2}^2 \big] + \frac{\sglip^2}{m^2} \sum_{j = 1}^{m} \Exs \big[ F (\theta_{t - 1}^{(j)}) - F (\thetastar) \big] \Big\},
  \end{align*}
  where $c_1 > 0$ is a universal constant.
\end{lemma}
\noindent See \Cref{subsubsec:proof-worker-async-bound} for the proof of \Cref{lemma:worker-async-bound}.

Taking these lemmas as given, let us now prove \Cref{thm:main-convex}. Substituting \Cref{lemma:main-recursion-cross-term-bound} and \Cref{lemma:main-recursion-variance-bound} into \Cref{eq:one-step-error-decomposition-in-thm-main-convex}, we have
\begin{multline*}
  \Exs \big[ \vecnorm{\theta_k - \thetastar}{2}^2 \big] \leq \big(1 - \strongconvex \stepsize_k / 2\big) \Exs \big[ \vecnorm{\theta_{k - 1} - \thetastar}{2}^2 \big] + \frac{c \stepsize_k^2}{m} \big( \sigstar^2 + (v^* + \Delta^2) \smallscale^2 \usedim \big) \\
  + \Big\{ c \stepsize_k^2 - \frac{\stepsize_k}{4 \smoothness} \Big\} \frac{1}{m} \sum_{j = 1}^{m} \Exs \big[ \vecnorm{\nabla F (\theta_{k - 1}^{(j)})}{2}^2 \big] \\
  + \Big\{c \frac{\sglip^2}{m}  \stepsize_k^2 - \frac{\stepsize_k}{2} \Big\} \frac{1}{m} \sum_{j = 1}^{m} \Exs \big[ F (\theta_{k - 1}^{(j)}) - F (\thetastar) \big]
  + 6 \smoothness \stepsize_k D_{k - 1}.
\end{multline*}
By taking stepsize satisfying the stability condition $\stepsize_k \leq \frac{c_0}{\smoothness + \sglip^2}$ for any $k \geq 1$, above inequality can be simplified to
\begin{multline*}
    \Exs \big[ \vecnorm{\theta_k - \thetastar}{2}^2 \big] \leq \exp \big( - \strongconvex \stepsize_k / 2\big) \Exs \big[ \vecnorm{\theta_{k - 1} - \thetastar}{2}^2 \big] + \frac{c \stepsize_k^2}{m} \big( \sigstar^2 + (v^* + \Delta^2) \smallscale^2 \usedim \big) + 6 \smoothness \stepsize_k  D_{k - 1} \\
    - \frac{\stepsize_k}{8 \smoothness m} \sum_{j = 1}^m  \Exs \big[ \vecnorm{\nabla F (\theta_{k - 1}^{(j)})}{2}^2 \big] - \frac{\stepsize_k}{4 m} \sum_{j = 1}^m \Exs \big[ F (\theta_{k - 1}^{(j)}) - F (\thetastar) \big]
    .
\end{multline*}
Define the accumulated time steps
\begin{align*}
  T (k) \mydefn \sum_{t = 1}^{k} \stepsize_t, \quad \mbox{for any } k \geq 1.
\end{align*}
Unrolling the recursion from $k = \tau_{i - 1}$ to $k = \tau_i$, we have
\begin{align*}
  \Exs \big[ \vecnorm{\theta_{\tau_i} - \thetastar}{2}^2 \big] &\leq e^{ - \tfrac{\strongconvex}{2} (T (\tau_i) - T (\tau_{i - 1}))} \Exs \big[ \vecnorm{\theta_{\tau_{i - 1} } - \thetastar}{2}^2 \big] + \sum_{k = \tau_{i - 1} + 1}^{\tau_i} e^{ - \tfrac{\strongconvex}{2} (T (\tau_i) - T (k))} \frac{c \stepsize_k^2}{m} \big( \sigstar^2 + (v^* + \Delta^2) \smallscale^2 \usedim \big)  \\
  &\qquad \qquad-  \frac{1}{4}\sum_{k = \tau_{i - 1} + 1}^{\tau_i} e^{ - \tfrac{\strongconvex}{2} (T (\tau_i) - T (k))} \frac{\stepsize_k}{m} \sum_{j = 1}^m \Big\{ \frac{1}{2 \smoothness} \Exs \big[ \vecnorm{\nabla F (\theta_{k - 1}^{(j)})}{2}^2 \big] +  \Exs \big[ F (\theta_{k - 1}^{(j)}) - F (\thetastar) \big] \Big\}  \\
  &\qquad \qquad \qquad \qquad+ 6 \smoothness  \sum_{k = \tau_{i - 1} + 1}^{\tau_i} e^{ - \tfrac{\strongconvex}{2} (T (\tau_i) - T (k))} \stepsize_k  D_{k - 1}.
\end{align*}
When the synchronization times satisfy $T (\tau_i) - T (\tau_{i - 1}) \leq 2 / \strongconvex$, above bound can be simplified as
\begin{align}
  \Exs \big[ \vecnorm{\theta_{\tau_i} - \thetastar}{2}^2 \big] &\leq e^{ - \tfrac{\strongconvex}{2} (T (\tau_i) - T (\tau_{i - 1}))} \Exs \big[ \vecnorm{\theta_{\tau_{i - 1} } - \thetastar}{2}^2 \big] + \sum_{k = \tau_{i - 1} + 1}^{\tau_i}  \frac{c \stepsize_k^2}{m} \big( \sigstar^2 + (v^* + \Delta^2) \smallscale^2 \usedim \big) \nonumber  \\
  &\qquad \qquad-  \frac{1}{4 e}\sum_{k = \tau_{i - 1} + 1}^{\tau_i}  \frac{\stepsize_k}{m} \sum_{j = 1}^m \Big\{ \frac{1}{2 \smoothness} \Exs \big[ \vecnorm{\nabla F (\theta_{k - 1}^{(j)})}{2}^2 \big] +  \Exs \big[ F (\theta_{k - 1}^{(j)}) - F (\thetastar) \big] \Big\} \nonumber \\
  &\qquad \qquad \qquad \qquad+ 6 \smoothness  \sum_{k = \tau_{i - 1} + 1}^{\tau_i} \stepsize_k  D_{k - 1}.\label{eq:scary-error-recursion-unroll-in-main-thm-convex-proof}
\end{align}
Invoking \Cref{lemma:worker-async-bound}, we note that
\begin{align*}
 & \sum_{k = \tau_{i - 1} + 1}^{\tau_i}  \stepsize_k  D_{k - 1} \\
  &\leq  c_1 ( v^* + \Delta^2)   \sum_{k = \tau_{i - 1} + 1}^{\tau_i} \stepsize_k  \sum_{t = \tau_{i - 1} + 1}^{k} \stepsize_t^2 \Big\{ \frac{\sigstar^2}{m} + \smallscale^2 \usedim +  \frac{1}{m} \sum_{j = 1}^m \Exs \big[ \vecnorm{\nabla F (\theta_{t - 1}^{(j)})}{2}^2 \big] + \frac{\sglip^2}{m^2} \sum_{j = 1}^{m} \Exs \big[ F (\theta_{t - 1}^{(j)}) - F (\thetastar) \big] \Big\}\\
  &\leq c_1 ( v^* + \Delta^2) \big( T(\tau_i) - T(\tau_{i - 1}) \big) \sum_{t = \tau_{i - 1} + 1}^{\tau_i} \stepsize_t^2 \Big\{ \frac{\sigstar^2}{m} + \smallscale^2 \usedim +  \frac{1}{m} \sum_{j = 1}^m  \Big( \Exs \big[ \vecnorm{\nabla F (\theta_{t - 1}^{(j)})}{2}^2 \big] + \frac{\sglip^2}{m} \Exs \big[ F (\theta_{t - 1}^{(j)}) - F (\thetastar) \big] \Big) \Big\}.
\end{align*}
When the synchronization times satisfy $T (\tau_i) - T (\tau_{i - 1}) < \frac{1}{2 \smoothness (v^* + \Delta^2)} $, and the stepsizes satisfy $\stepsize_t \leq \frac{1}{24e c_1 (\smoothness + \sglip^2)}$ for every $t$, the negative terms in \Cref{eq:scary-error-recursion-unroll-in-main-thm-convex-proof} can cancel the $ \Exs \big[ \vecnorm{\nabla F (\theta_{t - 1}^{(j)})}{2}^2 \big]$ and $\Exs \big[ F (\theta_{t - 1}^{(j)}) - F (\thetastar) \big]$ terms in the summation $\sum_{k = \tau_{i - 1} + 1}^{\tau_i}  \stepsize_k  D_{k - 1}$, leading to the recursive bound
\begin{align*}
  \Exs \big[ \vecnorm{\theta_{\tau_i} - \thetastar}{2}^2 \big] &\leq e^{ - \tfrac{\strongconvex}{2} (T (\tau_i) - T (\tau_{i - 1}))} \Exs \big[ \vecnorm{\theta_{\tau_{i - 1} } - \thetastar}{2}^2 \big] +  c \Big( \frac{\sigstar^2}{m} + (v^* + \Delta^2) \smallscale^2 \usedim \Big) \sum_{k = \tau_{i - 1} + 1}^{\tau_i}   \stepsize_k^2.
\end{align*}
Solving the recursion, we have
\begin{align*}
  \Exs \big[ \vecnorm{\theta_{\tau_r} - \thetastar}{2}^2 \big] \leq e^{- \frac{\strongconvex}{2} T (\tau_r)} \vecnorm{\theta_0 - \thetastar}{2}^2 + c e \Big( \frac{\sigstar^2}{m} + (v^* + \Delta^2) \smallscale^2 \usedim \Big)  \sum_{k = 1}^{\tau_r} e^{- \tfrac{\strongconvex}{2} (T (\tau_r) - T(k)) } \stepsize_k^2.
\end{align*}
For the summation term, we claim that
\begin{align}
  \sum_{k = 1}^{n} e^{- \tfrac{\strongconvex}{2} (T (n) - T(k)) } \stepsize_k^2 \leq 8 \stepsize_n, \label{eq:simple-recursion-bound-in-convex-proof}
\end{align}
whenever $(1 + \strongconvex \stepsize_{k} / 8) \stepsize_{k} \geq \stepsize_{k - 1}$ for every $k \geq 2$.

\paragraph{Proof of \Cref{eq:simple-recursion-bound-in-convex-proof}:} Define $A_n \mydefn \sum_{k = 1}^{n} e^{- \tfrac{\strongconvex}{2} (T (n) - T(k)) } \stepsize_k^2$. We prove the result by induction. For $n = 1$, clearly we have $A_1 = \stepsize_1^2 \leq 8 \stepsize_1$. Assuming that $A_{n - 1} \leq 8 \stepsize_{n - 1}$, for the case of $A_n$, we note the recursive formula
\begin{align*}
  A_n = e^{- \strongconvex \stepsize_n / 2} A_{n - 1} + \stepsize_n^2 \leq \big(1 - \strongconvex \stepsize_n / 4 \big) A_{n - 1} + \stepsize_n^2.
\end{align*}
By induction hypothesis, we have $A_{n - 1} \leq 8 \stepsize_{n - 1} / \strongconvex$. Substituting into the recursive bound, we have $A_n \leq 8 (1 - \strongconvex \stepsize_n / 4 ) \stepsize_{n - 1} / \strongconvex + \stepsize_n^2 \leq 8 (1 - \strongconvex \stepsize_n / 4 ) \stepsize_{n - 1} / \strongconvex + \stepsize_n^2 \leq 8 \tfrac{1 - \strongconvex \stepsize_n / 4 }{1 + \strongconvex \stepsize_n / 8} \stepsize_{n} / \strongconvex + \stepsize_n^2 \leq 8 \stepsize_n / \strongconvex$, which completes the induction proof.

\subsubsection{Proof of \Cref{lemma:main-recursion-cross-term-bound}}\label{subsubsec:proof-lemma-main-recursion-cross-term-bound}
We start by noting the following basic inequalities for strongly convex and smooth functions. For any $\theta \in \real^\usedim$, we have
\begin{subequations}
\begin{align}
  \inprod{\nabla F (\theta)}{\theta - \thetastar} &\geq \frac{\strongconvex \smoothness}{\strongconvex + \smoothness} \vecnorm{\theta - \thetastar}{2}^2 + \frac{1}{\strongconvex + \smoothness} \vecnorm{\nabla F (\theta)}{2}^2,\label{eq:basic-convex-property-1} \\
  \inprod{\nabla F (\theta)}{\theta - \thetastar} &\geq F (\theta) - F (\thetastar) + \frac{\strongconvex}{2} \vecnorm{\theta - \thetastar}{2}^2.\label{eq:basic-convex-property-2}
\end{align}
\end{subequations}

Applying \Cref{lemma:joint-quantization-channel} to the uplink transmission process in the $k$-th iteration, we can compute the conditional expectation.
\begin{align*}
   \Exs \big[ u_k \mid (\signalup_k^{(j)}, \beta_k^{(j)})_{j = 1}^m \big] &= \frac{1}{m} \sum_{j = 1}^m \nabla f (\theta_{k - 1}^{(j)}, X_k^{(j)}).
\end{align*}
Now we further take expectations conditionally on $\filtration_{k - 1}$ to obtain
$\Exs \big[ u_k \mid \filtration_{k - 1} \big] = \frac{1}{m} \sum_{j = 1}^m \nabla F (\theta_{k - 1}^{(j)})$, and therefore
\begin{align*}
  \Exs \big[ \inprod{\theta_{k - 1} - \thetastar}{u_k} \big] = \frac{1}{m} \sum_{j = 1}^m \Exs \Big[ \inprod{\theta_{k - 1} - \thetastar}{\nabla F (\theta_{k - 1}^{(j)})} \Big].
\end{align*}
For each $j \in [m]$, by applying \Cref{eq:basic-convex-property-1} and \Cref{eq:basic-convex-property-2}, we have
\begin{align*}
  \Exs \big[ \inprod{\theta_{k - 1}^{(j)} - \thetastar}{\nabla F (\theta_{k - 1}^{(j)})} \big] \geq  \frac{\strongconvex}{2} \Exs \big[ \vecnorm{\theta_{k - 1}^{(j)} - \thetastar}{2}^2 \big] + \frac{1}{2} \Exs \big[ F (\theta_{k - 1}^{(j)}) - F (\thetastar) \big] + \frac{1}{4 \smoothness} \Exs \big[ \vecnorm{\nabla F (\theta_{k - 1}^{(j)})}{2}^2 \big].
\end{align*}
By Cauchy--Schwarz inequality and Young's inequality, we also note that
\begin{align*}
  &\abss{\Exs \big[ \inprod{\theta_{k - 1}^{(j)} - \thetastar}{\nabla F (\theta_{k - 1}^{(j)})} \big] - \Exs \big[ \inprod{\theta_{k - 1} - \thetastar}{\nabla F (\theta_{k - 1}^{(j)})} \big]} \\
  & \leq \sqrt{\Exs \big[ \vecnorm{\theta_{k - 1}^{(j)} - \theta_{k - 1}}{2}^2 \big]} \cdot \sqrt{\Exs \big[ \vecnorm{\nabla F (\theta_{k - 1}^{(j)})}{2}^2 \big]} \leq \frac{1}{8 \smoothness} \Exs \big[ \vecnorm{\nabla F (\theta_{k - 1}^{(j)})}{2}^2 \big] + 2 \smoothness \Exs \big[\vecnorm{\theta_{k - 1}^{(j)} - \theta_{k - 1}}{2}^2 \big].
\end{align*}
Consequently, we can bound the inner product from below.
\begin{multline}
  \Exs \big[ \inprod{\theta_{k - 1} - \thetastar}{u_k} \big] \geq  \frac{\strongconvex}{2 m} \sum_{j = 1}^{m} \Exs \big[ \vecnorm{\theta_{k - 1}^{(j)} - \thetastar}{2}^2 \big] + \frac{1}{2 m} \sum_{j = 1}^m \Exs \big[ F (\theta_{k - 1}^{(j)}) - F (\thetastar) \big]\\
   + \frac{1}{8 \smoothness m} \sum_{j = 1}^m \Exs \big[ \vecnorm{\nabla F (\theta_{k - 1}^{(j)})}{2}^2 \big]
  - 2 \smoothness D_{k - 1}.\label{eq:main-lower-bound-in-cross-term-lemma}
\end{multline}
Finally, for the first term on the right-hand-side, we note that
\begin{align*}
  \Exs \big[ \vecnorm{\theta_{k - 1}^{(j)} - \thetastar}{2}^2 \big] \geq \frac{1}{2} \Exs \big[ \vecnorm{\theta_{k - 1} - \thetastar}{2}^2 \big] - \Exs \big[ \vecnorm{\theta_{k - 1}^{(j)} - \theta_{k - 1}}{2}^2 \big],
\end{align*}
for each $j \in [m]$. Averaging over $m$ workers, we have that
\begin{align*}
  \frac{\strongconvex}{2 m} \sum_{j = 1}^{m} \Exs \big[ \vecnorm{\theta_{k - 1}^{(j)} - \thetastar}{2}^2 \big] \geq \frac{\strongconvex}{4} \Exs \big[ \vecnorm{\theta_{k - 1}- \thetastar}{2}^2 \big] - \frac{\strongconvex}{2} D_{k - 1}.
\end{align*}
Substituting to \Cref{eq:main-lower-bound-in-cross-term-lemma} completes the proof of \Cref{lemma:main-recursion-cross-term-bound}.

\subsubsection{Proof of \Cref{lemma:main-recursion-variance-bound}}\label{subsubsec:proof-lemma-main-recursion-variance-bound}
By definition, we have $u_k = \frac{1}{m} \sum_{j = 1}^m A_\smallscale (\widehat{\signalup}_k^{(j)}, \beta_k^{(j)})$, where each $\widehat{\signalup}_k^{(j)}$ is obtained by applying the composed stochastic transformation $\correctionMap \circ \qtzadc \circ \channel \circ \qtzalg$ independently to the quantized signal $\Psi_\smallscale (\nabla f (\theta_{k - 1}^{(j)}, X_k^{(j)}))$. By \Cref{lemma:joint-quantization-channel}, we have
\begin{align*}
  \var \Big( A_\smallscale (\widehat{\signalup}_k^{(j)}, \beta_k^{(j)}) \mid \signalup_k^{(j)}, \beta_k^{(j)} \Big) \leq (4 v^* + \Delta^2) \big( 4 \vecnorm{\nabla f (\theta_{k - 1}^{(j)}, X_k^{(j)}) }{2}^2 + \smallscale^2 \usedim \big).
\end{align*}
Since the transmission between server and each workers are independent, we have
\begin{align*}
  \var \big( u_k \mid (\signalup_k^{(j)}, \beta_k^{(j)})_{j \in [m]} \big) \leq \frac{4 v^* + \Delta^2}{m} \Big\{ \smallscale^2 \usedim + \frac{4}{m} \sum_{j = 1}^m \vecnorm{\nabla f (\theta_{k - 1}^{(j)}, X_k^{(j)}) }{2}^2  \Big\}.
\end{align*}
Lemma~\ref{lemma:joint-quantization-channel} also implies that
\begin{align*}
  \Exs \big[ u_k \mid (\signalup_k^{(j)}, \beta_k^{(j)})_{j \in [m]} \big] = \frac{1}{m} \sum_{j = 1}^m \nabla f (\theta_{k - 1}^{(j)}, X_k^{(j)}) .
\end{align*}
We can then bound the total second moment as
\begin{align*}
  \Exs \big[ \vecnorm{u_k}{2}^2 \big] &\leq \Exs \Big[ \var \big( u_k \mid (\signalup_k^{(j)}, \beta_k^{(j)})_{j \in [m]} \big)\Big] + \Exs \Big\{ \vecnorm{\Exs \big[ u_k \mid (\signalup_k^{(j)}, \beta_k^{(j)})_{j \in [m]} \big]}{2} \Big\}^2\\
  &\leq  \frac{4 v^* + \Delta^2}{m} \Big\{ \smallscale^2 \usedim + \frac{4}{m} \sum_{j = 1}^m \Exs \big[ \vecnorm{\nabla f (\theta_{k - 1}^{(j)}, X_k^{(j)})  }{2}^2  \big]\Big\} + \Exs \Big[ \vecnorm{\frac{1}{m} \sum_{j = 1}^m \nabla f (\theta_{k - 1}^{(j)}, X_k^{(j)})}{2}^2 \Big].
\end{align*}
Conditionally on the filtration $\filtration_{k - 1}$, we can invoke \Cref{assume:stochastic-lipschitz} to obtain the bounds
\begin{align*}
  \Exs \big[ \vecnorm{\nabla f (\theta_{k - 1}^{(j)}, X_k^{(j)})  }{2}^2 \mid \filtration_{k - 1} \big] \leq  \vecnorm{\nabla F (\theta_{k - 1}^{(j)})}{2}^2 + \sigstarj{j}^2 + \sglip^2 \big( F (\theta_{k - 1}^{(j)}) - F (\thetastar) \big),
\end{align*}
and since the data points are independently sampled at each worker, we have
\begin{align*}
  &\Exs \Big[ \vecnorm{\frac{1}{m} \sum_{j = 1}^m \nabla f (\theta_{k - 1}^{(j)}, X_k^{(j)})}{2}^2 \mid \filtration_{k - 1} \Big]\\
  &= \vecnorm{\frac{1}{m} \sum_{j = 1}^m \nabla F (\theta_{k - 1}^{(j)})}{2}^2 + \frac{1}{m^2} \sum_{j = 1}^{m} \var \big( \nabla f (\theta_{k - 1}^{(j)}, X_k^{(j)}) \mid \filtration_{k - 1} \big)\\
  &\leq \frac{1}{m} \sum_{j = 1}^m \vecnorm{\nabla F (\theta_{k - 1}^{(j)})}{2}^2 + \frac{1}{m^2} \sum_{j = 1}^{m} \Big\{ \sigstarj{j}^2 + \sglip^2 \big( F (\theta_{k - 1}^{(j)}) - F (\thetastar) \big) \Big\}.
\end{align*}
Substituting these bounds into the variance bound, we have
\begin{multline*}
  \Exs \big[ \vecnorm{u_k}{2}^2 \big] \leq \frac{4 v^* + \Delta^2}{m} \smallscale^2 \usedim + \frac{1 + 16 v^* + 4 \Delta^2}{m} \Big\{ \sigstar^2 + \frac{\sglip^2}{m} \sum_{j = 1}^m \Exs [F (\theta_{k - 1}^{(j)}) -  F(\thetastar)] \Big\}\\
   + \Big\{1 + \frac{16 v^* + 4 \Delta^2}{m} \Big\} \frac{1}{m} \sum_{j = 1}^m \Exs \big[ \vecnorm{\nabla F (\theta_{k - 1}^{(j)})}{2}^2 \big],
\end{multline*}
which completes the proof of \Cref{lemma:main-recursion-variance-bound}.

\subsubsection{Proof of \Cref{lemma:worker-async-bound}}\label{subsubsec:proof-worker-async-bound}
Define the vector
\begin{align*}
  \widehat{u}_k^{(j)} \mydefn  A_\smallscale \big( \widehat{\signaldown}_k^{(j)}, \beta_k \big) = A_\smallscale \big( \correctionMap \circ \qtzadc \circ \channel \circ \qtzalg ( \Psi_\smallscale (u_k)), \beta_\smallscale (u_k) \big).
\end{align*}
The recursive update formulae can be written as
\begin{align*}
  \theta_k &= \theta_{k - 1} - \stepsize_k u_k,\\
  \theta_k^{(j)} &= \theta_{k - 1}^{(j)} - \stepsize_k  \widehat{u}_k^{(j)}, \quad \mbox{for } j \in [m].
\end{align*}
For each $j \in [m]$, we have the error expansion.
\begin{align*}
 \Exs \big[ \vecnorm{\theta_k - \theta_k^{(j)}}{2}^2 \big] = \Exs \big[\vecnorm{\theta_{k - 1} - \theta_{k - 1}^{(j)}}{2}^2 \big] + \stepsize_k^2 \Exs \big[ \vecnorm{u_k - \widehat{u}_k^{(j)}}{2}^2 \big] - 2 \stepsize_k \Exs \big[ \inprod{\theta_{k - 1} - \theta_{k - 1}^{(j)}}{u_k - \widehat{u}_k^{(j)}} \big].
\end{align*}
By \Cref{lemma:joint-quantization-channel}, we have
\begin{align*}
  \Exs \big[\widehat{u}_k^{(j)} \mid u_k \big] = u_k.
\end{align*}
Note that the transmission between server and workers in $k$-th round is independent of $\filtration_{k - 1}$. So we have
\begin{align*}
  \Exs \big[ \inprod{\theta_{k - 1} - \theta_{k - 1}^{(j)}}{u_k - \widehat{u}_k^{(j)}} \big] = 0.
\end{align*}
On the other hand, by \Cref{lemma:joint-quantization-channel}, we have the variance bound
\begin{align*}
  \Exs \big[ \vecnorm{u_k - \widehat{u}_k^{(j)}}{2}^2 \mid u_k \big] &\leq (4 v^* + \Delta^2) \big( 4 \vecnorm{u_k}{2}^2 + \smallscale^2 \usedim \big).
\end{align*}
Putting them together, we have the recursion
\begin{align}
  D_k \leq D_{k - 1} + \stepsize_k^2 \Exs \big[ \vecnorm{u_k - \widehat{u}_k^{(j)}}{2}^2 \big]
  \leq D_{k - 1} + \stepsize_k^2 (4 v^* + \Delta^2) \big( 4 \Exs \big[ \vecnorm{u_k}{2}^2 \big] + \smallscale^2 \usedim \big).\label{eq:Dk-recursive-bound}
\end{align}
On the other hand, when $k \in \{\tau_1, \tau_2, \cdots\}$, we have $\theta_k^{(j)} = \theta_k$ for all $j \in [m]$, and so $D_k = 0$.

For $k \in [\tau_{i - 1}, \tau_i)$, unrolling the recursion~\eqref{eq:Dk-recursive-bound} from $k$ down to $\tau_{i - 1}$, we have
\begin{align*}
  D_k \leq (4 v^* + \Delta^2)  \sum_{t = \tau_{i - 1} + 1}^{k} \stepsize_t^2 \big( 4 \Exs \big[ \vecnorm{u_t}{2}^2 \big] + \smallscale^2 \usedim \big).
\end{align*}
Substituting the bounds in \Cref{lemma:main-recursion-variance-bound}, we have
\begin{align*}
  D_k \leq c_1 ( v^* + \Delta^2)  \sum_{t = \tau_{i - 1} + 1}^{k} \stepsize_t^2 \Big\{ \frac{\sigstar^2}{m} + \smallscale^2 \usedim +  \frac{c}{m} \sum_{j = 1}^m \Exs \big[ \vecnorm{\nabla F (\theta_{t - 1}^{(j)})}{2}^2 \big] + c \frac{\sglip^2}{m^2} \sum_{j = 1}^{m} \Exs \big[ F (\theta_{t - 1}^{(j)}) - F (\thetastar) \big] \Big\},
\end{align*}
which completes the proof of \Cref{lemma:worker-async-bound}.

\subsection{Proof of \Cref{thm:non-convex}}\label{subsec:proof-thm-nonconvex}
By smoothness of the function $F$, we have
\begin{align}
  \Exs \big[F (\theta_k) \big] = \Exs \big[F (\theta_{k - 1}) \big] - \stepsize_k \Exs \big[ \inprod{\nabla F (\theta_{k - 1})}{u_k} \big] + \frac{\stepsize_k^2}{2} \smoothness \Exs [\vecnorm{u_k}{2}^2].\label{eq:non-convex-recursion}
\end{align}
We use the following lemma to control the cross term
\begin{lemma}\label{lemma:non-convex-cross-term}
  Under the setup of \Cref{thm:non-convex}, for each $k$, we have
  \begin{align*}
    \Exs \big[ \inprod{\nabla F (\theta_{k - 1})}{u_k} \big] \geq \frac{1}{4} \Exs \big[ \vecnorm{\nabla F (\theta_{k - 1})}{2}^2 \big] +  \frac{1}{4m} \sum_{j = 1}^m \Exs \big[ \vecnorm{\nabla F (\theta_{k - 1}^{(j)})}{2}^2 \big] - \frac{\smoothness^2}{2} D_{k - 1}.
  \end{align*}
\end{lemma}
\noindent See \Cref{subsubsec:proof-lemma-non-convex-cross-term} for the proof of this lemma.

Following the arguments in \Cref{lemma:main-recursion-variance-bound} using \Cref{assume:non-convex-sglip}, it is easy to see that
\begin{align}
  \Exs [\vecnorm{u_k}{2}^2]  \leq c \frac{\sigstar^2}{m} + c \frac{v^* + \Delta^2}{m} \smallscale^2 \usedim + \frac{c (1 + \lambda)}{m} \sum_{j = 1}^m \Exs \big[ \vecnorm{\nabla F (\theta_{k - 1}^{(j)})}{2}^2 \big].\label{eq:var-bound-non-convex}
\end{align}
Applying this bound to the arguments in \Cref{lemma:worker-async-bound}, it is easy to show that
\begin{align}
  D_k \leq   c_1 ( v^* + \Delta^2)  \sum_{t = \tau_{i - 1} + 1}^{k} \stepsize_t^2 \Big\{ \frac{\sigstar^2}{m} + \smallscale^2 \usedim +  \frac{1 + \lambda}{m} \sum_{j = 1}^m \Exs \big[ \vecnorm{\nabla F (\theta_{t - 1}^{(j)})}{2}^2 \big] \Big\},\label{eq:Dk-bound-non-convex}
\end{align}
for each $k \in [\tau_{i - 1}, \tau_i)$.

Now we substitute \Cref{lemma:non-convex-cross-term} and \Cref{eq:var-bound-non-convex} into \Cref{eq:non-convex-recursion}. Telescoping the summation from time $0$ to time $n$, we note that
\begin{multline}
  \frac{1}{4} \sum_{k = 1}^n \stepsize_k \Exs \big[ \vecnorm{\nabla F (\theta_{k - 1})}{2}^2 \big] + \frac{1}{4 m} \sum_{k = 1}^n \stepsize_k \sum_{j = 1}^{m} \Exs \big[ \vecnorm{\nabla F (\theta_{k - 1}^{(j)})}{2}^2 \big] - \frac{\smoothness^2}{2} \sum_{k = 1}^n \stepsize_k D_{k - 1}\\
   \leq \Exs [F (\theta_0)] - \Exs [ F(\theta_n)] + \sum_{k = 1}^n \stepsize_k^2 \smoothness \Big\{  c \frac{\sigstar^2}{m} + c \frac{v^* + \Delta^2}{m} \smallscale^2 \usedim + \frac{c (1 + \lambda)}{m} \sum_{j = 1}^m \Exs \big[ \vecnorm{\nabla F (\theta_{k - 1}^{(j)})}{2}^2 \big] \Big\}.\label{eq:scary-recursion-non-convex}
\end{multline}
By \Cref{eq:Dk-bound-non-convex}, if the synchronization times satisfy $T (\tau_i) - T (\tau_{i - 1}) \leq 1 / (2 \smoothness)$ for each $i = 1,2,\cdots$, we have
\begin{align*}
  \sum_{k = 1}^n \stepsize_k D_{k - 1} \leq \frac{c_1 ( v^* + \Delta^2)}{2 \smoothness}  \sum_{t = 1}^n \stepsize_t^2 \Big\{ \frac{\sigstar^2}{m} + \smallscale^2 \usedim +  \frac{1 + \lambda}{m} \sum_{j = 1}^m \Exs \big[ \vecnorm{\nabla F (\theta_{t - 1}^{(j)})}{2}^2 \big] \Big\}
\end{align*}
Substituting back to \Cref{eq:scary-recursion-non-convex}, when the stepsize satisfies $\stepsize_k \leq \frac{c_0}{\smoothness (1 + \lambda)}$, we have
\begin{align*}
  \frac{1}{4} \sum_{k = 1}^n \stepsize_k \Exs \big[ \vecnorm{\nabla F (\theta_{k - 1})}{2}^2 \big] 
  \leq \Exs [F (\theta_0)] - \Exs [ F(\theta_n)] +  c\smoothness \Big\{  \frac{\sigstar^2}{m} +  (v^* + \Delta^2) \smallscale^2 \usedim \Big\} \cdot \sum_{k = 1}^n \stepsize_k^2.
\end{align*}
By the definition of the random variable $R$, we have
\begin{align*}
  \sum_{k = 1}^n \stepsize_k \Exs \big[ \vecnorm{\nabla F (\theta_{k - 1})}{2}^2 \big] = \Exs \big[ \vecnorm{\nabla F (\theta_R)}{2}^2  \big] \cdot \sum_{k = 1}^n \stepsize_k.
\end{align*}
Substituting back to the telescope formula completes the proof.

\subsubsection{Proof of \Cref{lemma:non-convex-cross-term}}\label{subsubsec:proof-lemma-non-convex-cross-term}

By \Cref{lemma:joint-quantization-channel}, we have $\Exs \big[ u_k \mid \filtration_{k - 1} \big] = \frac{1}{m} \sum_{j = 1}^m \nabla F (\theta_{k - 1}^{(j)})$, and consequently
\begin{align*}
  &\Exs \big[ \inprod{\nabla F (\theta_{k - 1})}{u_k} \big] = \frac{1}{m} \sum_{j = 1}^{m}\Exs \big[ \inprod{\nabla F (\theta_{k - 1})}{\nabla F (\theta_{k - 1}^{(j)})} \big]  \\
  &\geq \Exs \big[ \vecnorm{\nabla  F (\theta_{k - 1})}{2}^2 \big] - \frac{1}{m} \sum_{j = 1}^m \sqrt{\Exs \big[ \vecnorm{\nabla  F (\theta_{k - 1})}{2}^2 \big]} \cdot \sqrt{\Exs \big[ \vecnorm{\nabla F(\theta_{k - 1}^{(j)}) - \nabla  F (\theta_{k - 1})}{2}^2 \big]}  \\
  &\geq \frac{1}{2} \Exs \big[ \vecnorm{\nabla  F (\theta_{k - 1})}{2}^2 \big] - \frac{\smoothness^2}{2} D_{k - 1},
\end{align*}
where in the last inequality, we use the smoothness of the function $F$ and the Cauchy--Schwarz inequality.

On the other hand, we note that
\begin{align*}
  &\Exs \big[ \inprod{\nabla F (\theta_{k - 1})}{u_k} \big] = \frac{1}{m} \sum_{j = 1}^{m}\Exs \big[ \inprod{\nabla F (\theta_{k - 1})}{\nabla F (\theta_{k - 1}^{(j)})} \big] \\
  &\geq  \frac{1}{m} \sum_{j = 1}^{m}\Exs \big[ \vecnorm{\nabla F (\theta_{k - 1}^{(j)})}{2}^2 \big] - \frac{1}{m} \sum_{j = 1}^m \sqrt{\Exs \big[ \vecnorm{\nabla  F (\theta_{k - 1}^{(j)})}{2}^2 \big]} \cdot \sqrt{\Exs \big[ \vecnorm{\nabla F(\theta_{k - 1}^{(j)}) - \nabla  F (\theta_{k - 1})}{2}^2 \big]} \\
  &\geq \frac{1}{2 m} \sum_{j = 1}^{m}\Exs \big[ \vecnorm{\nabla F (\theta_{k - 1}^{(j)})}{2}^2 \big] - \frac{\smoothness^2}{2} D_{k - 1}.
\end{align*}
Combining the two bounds completes the proof of this lemma.

\section{Discussion}\label{sec:discussion}
In this paper, we propose a novel algorithmic framework for communication-efficient distributed learning with quantized gradients over bi-directional noisy channels. We introduce three key technical components:
\begin{itemize}
  \item A post-coding scheme that ensures unbiasedness of the transmitted signal, even with non-linear quantization.
  \item A scale-adaptive transformation that dynamically adjusts the quantization levels.
  \item A federated learning framework that stabilizes the training process by synchronizing model parameters at certain frequencies.
\end{itemize}
Our theoretical results demonstrate that, under standard assumptions on the loss function and stochastic gradients, the proposed method achieves convergence rates comparable to those of fully centralized methods, even in the presence of low-resolution ADC/DAC quantization and high channel noise. We also provide empirical evidence supporting our theoretical findings: the proposed scheme achieves the same test accuracy with less than 20\% of the communication cost compared to fully coded communications. The simulation results further illustrate that all the technical components are indispensable for achieving the desired performance.

This work opens several avenues for future research. One promising direction is to explore improved schemes to transmit the coded part in scale-adaptive transformation, which could further reduce the communication overhead. Additionally, investigating the impact of different network topologies on the performance of the proposed framework could yield valuable insights. Finally, extending the algorithms to accommodate more realistic communication channel models and hardware constraints remains an important challenge.

\section*{Acknowledgments}
This work was partially supported by NSERC grant RGPIN-2024-05092 and a Connaught New Researcher Award to WM. The authors thank Ruocheng Wang at Marvell for helpful discussions.

\bibliographystyle{alpha}
\bibliography{references}

\end{document}

%% file: final_macros.tex
\newcommand{\stepsize}{\eta}

\newcommand{\real}{\ensuremath{\mathbb{R}}}

\newcommand{\thetastar}{\ensuremath{\theta^*}}

\newcommand{\abss}[1]{\left| #1 \right |}

\newcommand{\mydefn}{\ensuremath{:=}}

\newcommand{\smoothness}{L} 

\newcommand{\matsnorm}[2]{|\!|\!| #1 | \! | \!|_{{#2}}}
\newcommand{\vecnorm}[2]{\| #1\|_{#2}}

\newcommand{\inprod}[2]{\ensuremath{\langle #1 , \, #2 \rangle}}

\newcommand{\Exs}{\ensuremath{{\mathbb{E}}}}
\newcommand{\Prob}{\ensuremath{{\mathbb{P}}}}

\DeclareMathOperator{\var}{var}

\newtheoremstyle{named}{}{}{\itshape}{}{\bfseries}{.}{.5em}{\thmnote{#3's }#1}
\theoremstyle{named}

\theoremstyle{plain}

\newtheorem{theorem}{Theorem}
\newtheorem{proposition}{Proposition}

\newtheorem{lemma}{Lemma}

\newlength{\widebarargwidth}
\newlength{\widebarargheight}
\newlength{\widebarargdepth}

\makeatletter
\long\def\@makecaption#1#2{
        \vskip 0.8ex
        \setbox\@tempboxa\hbox{\small {\bf #1:} #2}
        \parindent 1.5em  
        \dimen0=\hsize
        \advance\dimen0 by -3em
        \ifdim \wd\@tempboxa >\dimen0
                \hbox to \hsize{
                        \parindent 0em
                        \hfil
                        \parbox{\dimen0}{\def\baselinestretch{0.96}\small
                                {\bf #1.} #2
                                }
                        \hfil}
        \else \hbox to \hsize{\hfil \box\@tempboxa \hfil}
        \fi
        }
\makeatother

\long\def\comment#1{}
\definecolor{battleshipgrey}{rgb}{0.52, 0.52, 0.51}
\definecolor{darkgray}{rgb}{0.66, 0.66, 0.66}
\definecolor{darkgreen}{rgb}{0.0, 0.2, 0.13}
\definecolor{darkspringgreen}{rgb}{0.09, 0.45, 0.27}
\definecolor{dukeblue}{rgb}{0.0, 0.0, 0.61}
\definecolor{olivedrab7}{rgb}{0.24, 0.2, 0.12}
\definecolor{darkblue}{rgb}{0.0, 0.0, 0.55}
\definecolor{darkscarlet}{rgb}{0.34, 0.01, 0.1}
\definecolor{candyapplered}{rgb}{1.0, 0.03, 0.0}
\definecolor{ao(english)}{rgb}{0.0, 0.5, 0.0}
\definecolor{applegreen}{rgb}{0.55, 0.71, 0.0}
